\documentclass{article} 

\usepackage[table, dvipsnames, x11names]{xcolor}
\usepackage[accepted]{icml2023}

\usepackage{hyperref}
\hypersetup{
    colorlinks,
    linkcolor={red!50!black},
    citecolor={blue!50!black},
    urlcolor={blue!80!black}
}

\usepackage{url}

\usepackage{amsmath,amssymb}

\usepackage{booktabs}
\usepackage{siunitx}
\usepackage{makecell}
\usepackage{booktabs}
\usepackage{multirow}

\usepackage{graphicx}
\usepackage{tabularx}
\usepackage{wrapfig}
\usepackage{pgfplots}
\usepackage{tikz}
\usepackage{subcaption}
\usepackage[font={footnotesize}]{caption} 

\usepackage{algorithm}
\usepackage{algorithmic}

\usepackage{textcomp}

\usepackage[nameinlink,capitalise]{cleveref}
\crefname{figure}{Figure}{Figures}
\Crefname{figure}{Figure}{Figures}

\usepackage[utf8]{inputenc}
\usepackage{mathtools, nccmath}

\usepackage{enumitem} 

\usepackage[dvipsnames]{xcolor}
\usepackage{soul}
\usepackage{lipsum}  

\usepackage{amsmath,amsfonts,bm}

\def\eqref#1{equation~\ref{#1}}

\def\1{\bm{1}}

\DeclareMathAlphabet{\mathsfit}{\encodingdefault}{\sfdefault}{m}{sl}
\SetMathAlphabet{\mathsfit}{bold}{\encodingdefault}{\sfdefault}{bx}{n}

\DeclareMathOperator*{\argmax}{arg\,max}

\newcommand{\ours}{Meta-SAGE}
\newcommand{\meta}{SML}
\newcommand{\adpt}{SAGE}

\def\xunderbrace#1_#2{{\underbrace{#1}_{#2}}}
\def\xoverbrace#1^#2{{\overbrace{#1}^{#2}}}

\definecolor{olive}{rgb}{0.6, 0.6, 0.2}
\definecolor{sand}{rgb}{0.8666666666666667, 0.8, 0.4666666666666667}
\definecolor{wine}{rgb}{0.5333333333333333, 0.13333333333333333, 0.3333333333333333}
\definecolor{deblue}{RGB}{11,132,147}
\definecolor{ocra}{RGB}{204, 119, 34}
\definecolor{darkgreen}{rgb}{0.0, 0.5, 0.0}
\definecolor{darkblue}{rgb}{0.0, 0.2, 0.55}

\def \papertitle{\ours{}: Scale Meta-Learning Scheduled Adaptation with Guided Exploration for Mitigating Scale Shift on Combinatorial Optimization}

\icmlsetsymbol{equal}{*}

\icmltitlerunning{\ours{}}
\begin{document}


\twocolumn[
\begin{icmlauthorlist}

\icmltitle{\papertitle}

\icmlauthor{Jiwoo Son}{kaist,equal} 
\icmlauthor{Minsu Kim}{kaist,equal} 
\icmlauthor{Hyeonah Kim}{kaist} 
\icmlauthor{Jinkyoo Park}{kaist}

\end{icmlauthorlist}

\icmlaffiliation{kaist}{Department of Industrial and System Engineering, KAIST, Korea}


\icmlcorrespondingauthor{Jinkyoo Park}{jinkyoo.park@kaist.ac.kr}

\icmlkeywords{Machine Learning, ICML}

\vskip 0.3in
]


\printAffiliationsAndNotice{\icmlEqualContribution} 
\sethlcolor {Aquamarine}
\begin{abstract}

This paper proposes \ours{}, a novel approach for improving the scalability of deep reinforcement learning models for combinatorial optimization (CO) tasks. Our method adapts pre-trained models to larger-scale problems in test time by suggesting two components: a scale meta-learner (SML) and scheduled adaptation with guided exploration (SAGE). First, SML transforms the context embedding for subsequent adaptation of SAGE based on scale information. Then, SAGE adjusts the model parameters dedicated to the context embedding for a specific instance. 
SAGE introduces locality bias, which encourages selecting nearby locations to determine the next location. The locality bias gradually decays as the model is adapted to the target instance.
Results show that \ours{} outperforms previous adaptation methods and significantly improves scalability in representative CO tasks.
Our source code is available at \href{https://github.com/kaist-silab/meta-sage}{https://github.com/kaist-silab/meta-sage}. 

\end{abstract}

\section{Introduction}

Combinatorial optimization (CO) is a task for finding an optimal combination of discrete variables which contain important problems, e.g., the traveling salesman problem (TSP): finding the shortest path of the Hamiltonian cycle. Solving CO problems is crucial because it can be applied to several high-impact tasks such as logistics \citep{transportation} and DNA sequencing \citep{caserta2014hybrid}. However, solving CO is usually NP-hard, and it is challenging to design an exact solver practically. 
Thus, hand-crafted heuristic solvers have been widely used to generate reasonable solutions fast \citep{concorde, lkh2017}. 
Despite of their practicality, heuristic solvers are designed based on problem-specific properties, so they cannot solve other kinds of problems in general (i.e., limited expandability to other problems). Even if it is possible, applying the heuristic solvers to other problems requires domain knowledge.

Deep learning approaches have recently been used to tackle the limited expandability of heuristic solvers by automating their design process. There are two approaches according to solving strategies: \textit{constructive} and \textit{improvement} methods. Constructive methods start from the empty solution and sequentially assign a decision variable to construct a complete solution \citep[e.g.,][]{pointer, kool2018attention}. On the other hand, improvement methods start with a complete solution and iteratively revise the given solution to improve solutions \citep[e.g.,][]{NLNS, chen2019learning}.
Also, deep learning algorithms for CO can be categorized according to learning strategies: \textit{supervised learning} \cite{pointer,xin2021neurolkh,li2021learning,hottung2020learning,kool_dp,fu2020generalize,li2021learning} and \textit{deep reinforcement learning} (DRL) \cite{khalil,bello2017neural,li2018combinatorial,duedon,Nazari,NLNS,kool2018attention,chen2019learning,ma2019combinatorial,drl-2opt,kwon2020pomo,barrett2020exploratory,wu2020learning,mis,xin2021multi,park2021schedulenet,park2021learning,yoon2021deep,kwon2021matrix,kim2021learning,kim2022sym,ma2022efficient,qiu2022dimes}. In this work, we focus on the DRL constructive methods\footnote{A model or policy refers to a DRL constructive method unless there is a specific description in the rest of this paper.}, which are beneficial to solve different kinds of problems since (1) the constructive method can effectively satisfy the constraint (e.g., capacity limit of the vehicle) using masking scheme \citep{bello2017neural,kool2018attention}, (2) DRL can generate a solver without labeled data from expert-level heuristics.

One of the critical challenges of the DRL method is scalability. There are two major directions to mitigate scalability. First, zero-shot contextual methods try to capture the contextual feature of each instance \citep{kool2018attention,mis,kwon2020pomo,park2021schedulenet,qiu2022dimes,kim2022sym}. They train policies to utilize these context embeddings (i.e., contextual multi-task learning) and make zero-shot inferences for new problems. 
\citet{mis} and \citet{park2021schedulenet} try to overcome scalability issues by effectively capturing inter-relationship between nodes via graph representation learning. While their methods give good feasibility of zero-shot solving of general CO problems, the performances are not competitive compared to effective heuristics.

On the other hand, \citet{bello2017neural,hottung2021efficient,choo2022simulation}  suggested an effective transfer learning scheme that adapts the DRL model pre-trained on small-scale problems to solve a larger-scale problem. 
The scheme is a test time adaptation that directly solves target problems by iteratively solving the target problem and modifying the parameters of the model. Adaptation transfer learning improves scalability with high performance. However, when the distributional shift of scale between the source and target problems becomes large, it requires massive iterations to properly update the model's parameters, leading to inefficiency at the test time.

\textbf{Contribution.} 
This paper improves transferability to a much larger scale by suggesting Meta-SAGE, which combines contextual meta-learning and parameter adaptation. Our target is to iterative adapt the DRL model pre-trained with small-scale CO problems to the larger-scale CO problem in the test-time, the same as \citet{hottung2021efficient} and \citet{choo2022simulation}. 
The transferability is quantified by the level of reducing the number of adaptation iterations $K$ or (2) improving performance on a fixed number of iterations $K=K'$. To achieve our goal, we propose two novel components, a scale meta-learner (SML) and Scheduled adaptation with guided exploration (SAGE):

\begin{itemize}
\item \textbf{A scale meta-learner (SML)} generates scale-aware context embedding by considering the subsequent parameter updates of SAGE, i.e., amortizing the parameter update procedure in training. SML is meta-learned from the bi-level structure that contains SAGE operation as a lower-level problem.

\item \textbf{Scheduled adaptation with guided exploration (SAGE)} efficiently updates parameters to adapt to the target instance at the test time. SAGE introduces locality bias, which encourages selecting nearby locations to determine the next visit. The locality bias gradually decays as the model is adapted to the target.
\end{itemize}

The \ours{} performs better than existing DRL-based CO model adaptation methods at four representative CO tasks: TSP, the capacitated vehicle routing problem (CVRP), the prize collecting TSP, and the orienteering problem. For example, \ours{} gives $0.68\%$ optimal gap at the CVRP scale of $1,000$, whereas the current state-of-the-art adaptation method \citep{hottung2020learning} gives $3.95\%$ gap. 

Notably, \ours{} outperforms the representative problem-specific heuristic on some tasks.

\section{Preliminary and Related Works}

This section provides preliminaries for DRL-based 
methods for combinatorial optimization and transfer learning-based adaptation schemes.  

\subsection{Constructive DRL methods for CO}

\begin{figure}
  \centering
  \includegraphics[width=1.0\linewidth]{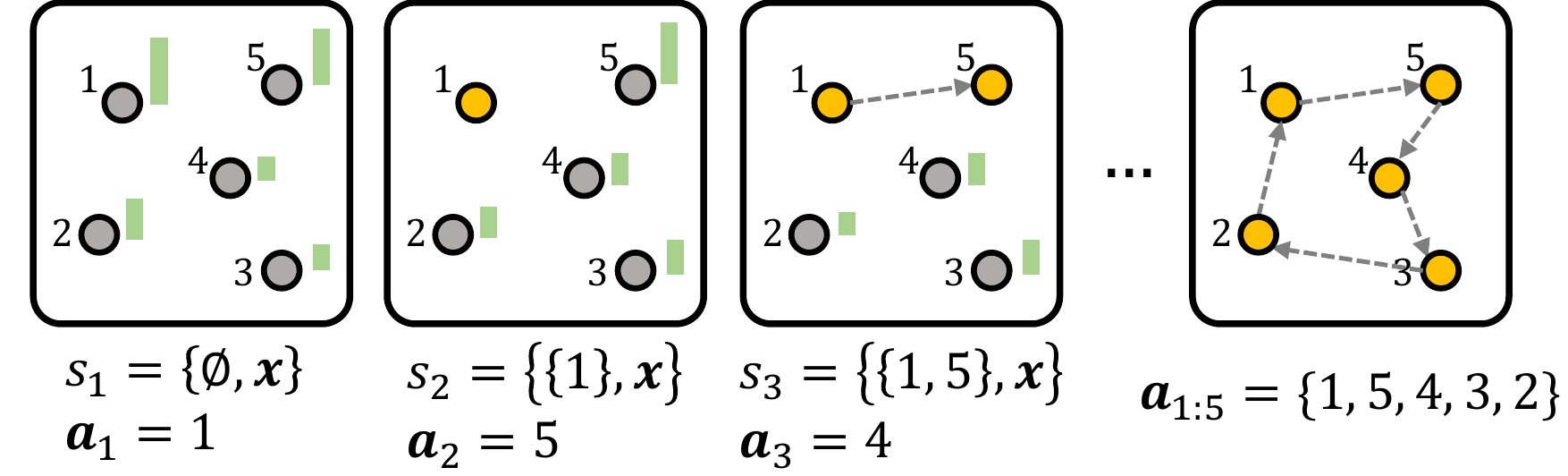}
  \caption{An example: let us have five locations to visit $\boldsymbol{x}=\{x_1,x_2,x_3,x_4, x_5\}$. The policy starts with the empty solution $\emptyset$, and constructs the solution by adding the next visit at each step based on the partial solution $\bm{a}_{1:t-1}$.}
  \label{fig:tsp_ex}
\end{figure}

An example of a constructive DRL method for TSP (other CO problems can be formulated similarly) is as follows:
\begin{itemize}
    \item States: a state $s_t$ is defined as a problem $\boldsymbol{x}=\{x_i\}_{i=1}^{N}$ that contains $N$ locations to visit and a sequence of previous actions $\boldsymbol{a}_{1:t-1}$, i.e., $\boldsymbol{s}_t = \{\boldsymbol{a}_{1:t-1};\boldsymbol{x}\}$.
    \item Actions: An action $\boldsymbol{a}_t$ is selecting the next visit out of unvisited locations $\{1,...,N\}\setminus \boldsymbol{a}_{1:t-1}$.
    \item Reward: the reward R at the final state is the negative tour length of a complete solution, i.e., $R(\bm{a}_{1:N}, \bm{x}) = -\sum_{i=1}^{N}||x_{\boldsymbol{a}_{i+1}} - x_{\boldsymbol{a}_{i}}||_{2}$, where $x_{\bm{a}_{i}}$ is the 2D coordination selected by action $\bm{a}_{i}$.
    \item Policy: The policy for the complete solution for problem $\boldsymbol{x}$ is $p_\theta(\boldsymbol{a}_{1:N}|\boldsymbol{x}) = \prod_{t=1}^N p_\theta(\boldsymbol{a}_t|\boldsymbol{s}_t)$, where $p_\theta(\boldsymbol{a}_t|\boldsymbol{s}_t)$ is the probability of selecting action  $\boldsymbol{a}_t$ given state $\boldsymbol{s}_t$.
\end{itemize}

As illustrated in \cref{fig:tsp_ex}, the policy constructs a complete solution starting with an empty solution.
This MDP is contextualized by $\boldsymbol{x}$ as the policy and reward highly rely on problem $\boldsymbol{x}$; thus, the trained policy differs according to the problem distribution $P(\boldsymbol{x}) = P(\boldsymbol{x}|N)P(N)$.
Formally, the DRL method aims to find parameter $\theta$ such that:
\begin{equation}
    \theta^{*} = \argmax_{\theta} \mathbb{E}_{P(N)}\mathbb{E}_{P(\boldsymbol{x}|N)}\mathbb{E}_{p_{\theta}(\cdot|\boldsymbol{x})}\left[R(\boldsymbol{a}_{1:N},\boldsymbol{x})\right]
    \label{eq:drl_training}
\end{equation}
Because reward function $R(\cdot)$ is non-differentiable, the REINFORCE-based method with a proper baseline for reducing variances has been utilized \citep{bello2017neural,kool2018attention,kwon2020pomo,kim2022sym}. As shown in \cref{eq:drl_training}, the distribution $P(\bm{x})$ would affect the learned policy $p_\theta$.
In the previous literature \cite{kool2018attention}, $P(N)$ is not considered while fixing $N$ to be constant, and $P(\boldsymbol{x}|N)$ was assumed as a uniform distribution, which causes the distribution shift issue when the policy is used for solving different size of the problem (i.e., $N'>N$).

\subsection{Transfer Learning-based Adaptation Methods for DRL-CO}

\textbf{Active Search \citep[AS; ][]{bello2017neural}} is a transfer learning method that updates the pre-trained parameters during the inference of a solution for a specific instance $\boldsymbol{x}$. Usually, the DRL model is trained to maximize rewards on the problem distribution $P(\boldsymbol{x}|N)$.
Therefore, the trained model can solve an arbitrary instance $\bm{x}$ sampled from $P(\boldsymbol{x}|N)$ at the test time.
However, active search improves model performance while focusing on a single instance $\boldsymbol{x}$. While active search gives promising performances, updating parameters in the test time is often inefficient and degrades practicality.

\textbf{Efficient Active Search \citep[EAS; ][]{hottung2021efficient}} is an improved version of active search that updates a subset of parameters, not whole parameters. EAS adds new parameters dedicated to embedding and updates these parameters for the specific target problem while maintaining the rest of the parameters. This separation of the parameters' roles allows an efficient update.

However, these methods transfer the pre-trained parameters to target problem $\boldsymbol{x}$ without utilizing prior information of distributional shift, e.g., scale gap from source to target. We observe that giving the scale priors to the transfer learning can improve transferability.

\begin{figure*}
\centering
\begin{minipage}[b]{0.49\textwidth}
  \includegraphics[width=\textwidth]{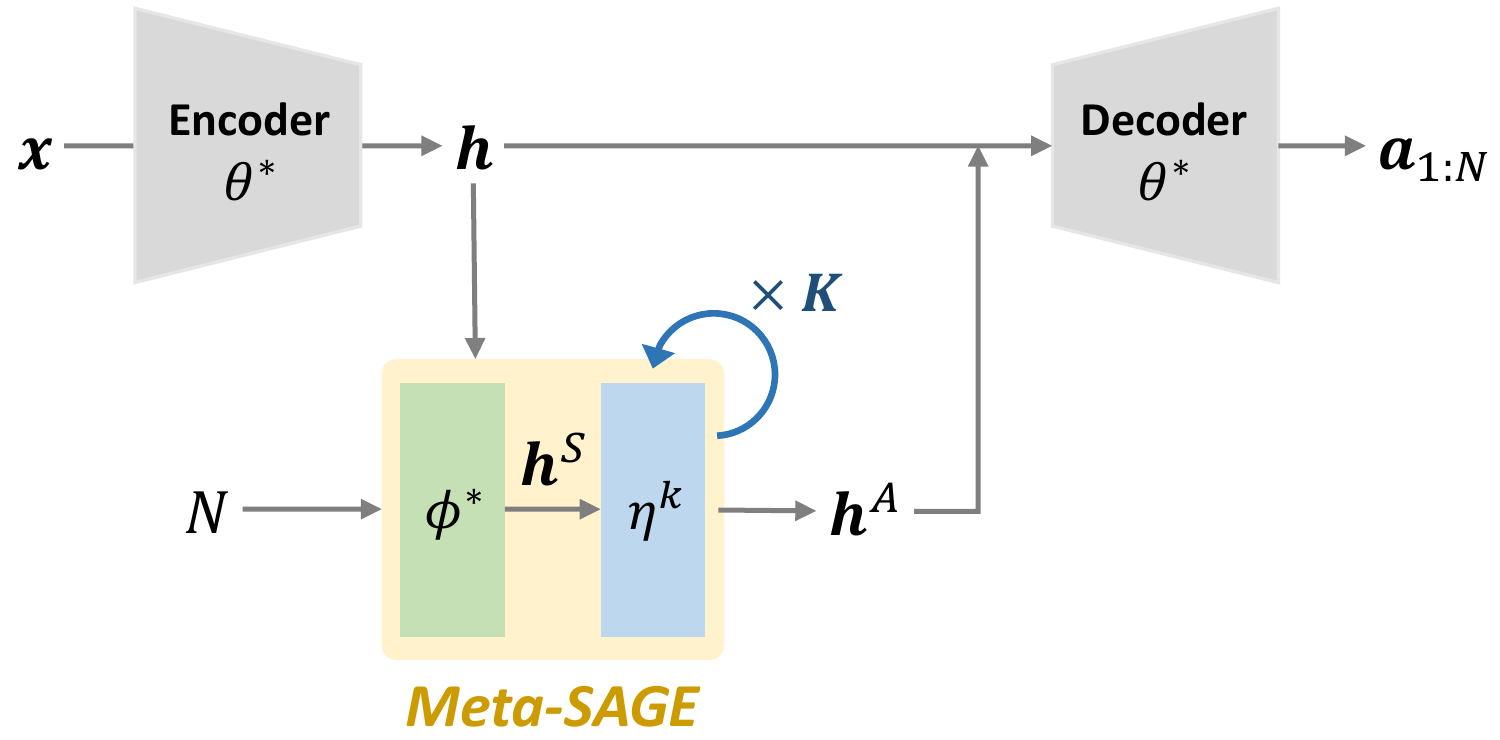}
  \caption{Structure of \ours{} and test time adaptation. The scale-conditioned initial embedding $\bm{h}^S$ is obtained by $\textcolor{darkgreen}{g_{\phi^*}}(\bm{x}, N)$ and the instance-wise target embedding $\bm{h}^A$ is obtained by $\textcolor{darkblue}{g_{\eta^{(k)}}}(\bm{h}^S)$. At $k$-th adaptation, $\eta^{k}$ is updated for the test instance $\bm{x}$, and other parameters (i.e., $\theta^*$ and $\phi^*$) are fixed.
  }
  \label{fig:structure}
\end{minipage}
\hfill
  \centering
\begin{minipage}[b]{0.48\textwidth}
  \includegraphics[width=\linewidth]{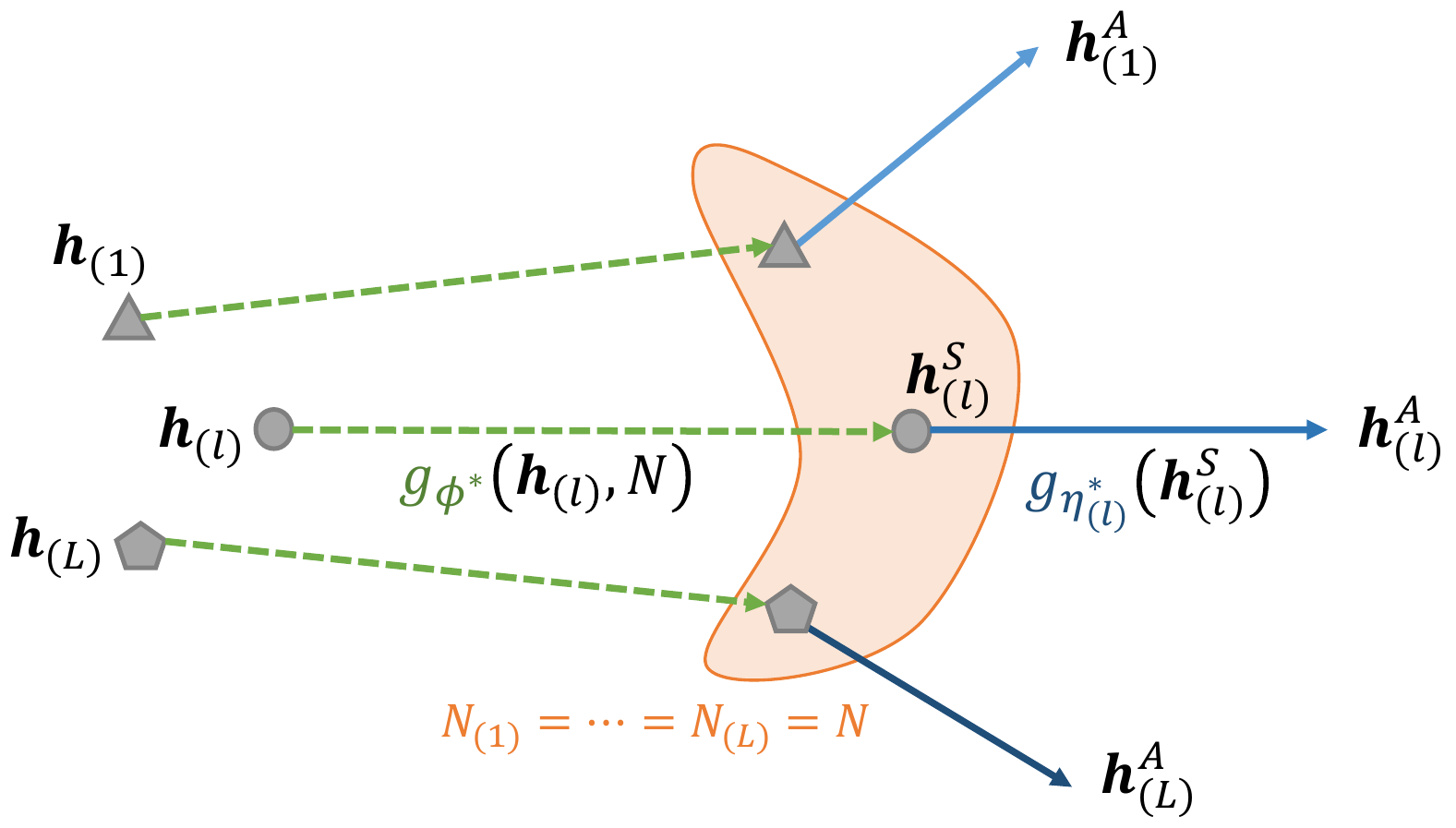}
  \caption{Transformation of embedding $\bm{h}$. 
  SML transforms the original embedding $\bm{h}_{(l)}$ to the scale-conditioned initial embedding $\bm{h}^S_{(l)} = g_{\phi^*}(h_{(l)}, N)$, where $N=N_{(l)}$. 
  Then, $\eta^*_{(l)}$ is obtained via the SAGE operation with $K$ iteration for each instance. 
  Finally, MLP $g_{\eta^*_{(l)}}(\bm{h}^S_{(l)})$ provides the target embedding $\bm{h}^A_{(l)}$.
  }
  \label{fig:meta}
\end{minipage}
\end{figure*}

\section{Adaptation with \ours{}} \label{sec:method}

This section explains how to adapt pre-trained models with \ours{}, which consists of a scale meta-learner (\meta{}) and scheduled adaptation with guided exploration (\adpt{}).
As shown in \cref{fig:structure}, new parameters $\phi$ and $\eta$ are introduced for \meta{} and \adpt{}, respectively.
The multi-layer perceptron (MLP) layers $g_{\phi}(\cdot)$ and $g_{\eta}(\cdot)$ transform the the original context embedding $\bm{h}$ based on the target scale $N$, i.e., $\bm{h}^A = g_{\eta}(g_{\phi} (\bm{h}, N))$.
Note that $\phi$ is pre-trained to capture the scale information, and $\eta$ is solely updated at test time adaptation. 
This section focus on explaining adaptation procedure of \ours{}; see \cref{sec:learning} for the learning scheme for \meta{}.

\subsection{Decision Process and Guided Exploration}

\paragraph{Decision process of DRL models.} 
We focus on the models utilizing the encoder-decoder structure \citep{pointer,bello2017neural,kool2018attention,kwon2020pomo,kim2022sym} where the encoder captures context embedding $\boldsymbol{h}=\{h_i\}_{i=1}^{N}$ of the problem $\boldsymbol{x}=\{x_i\}_{i=1}^{N}$, and the decoder sequentially generates action $\boldsymbol{a}_t$ with the context embedding $\boldsymbol{h}$ and the partial solution $\boldsymbol{a}_{1:t-1}$.

At each decision step $t$, the decoder computes the compatibility vector $\bm{u}$ to be used to select the next action. There are several strategies to compute compatibility vectors; the most popular way is to utilize multi-head attention \citep{transformer,kool2018attention,kwon2020pomo,kim2022sym}. To mask visited locations, the compatibility vector $\bm{u}$, is expressed as:

\begin{equation*}
    u_j = \begin{cases}
    u_j & \mbox{ if } j \notin \bm{a}_{1:t-1} \\
    - \infty & \mbox{ otherwise.}
    \end{cases}
\end{equation*}

The probability $p_i$ for selecting location $i$ is computed by taking a SoftMax with temperature $\mathcal{T}$ as follows:
\begin{equation} \label{eq:policy_softmax}
    p_i = \frac{e^{u_i / \mathcal{T}}}{\sum_j e^{u_j / \mathcal{T}}} \quad \forall i=1, \ldots, N.
\end{equation}
We provide detailed implementation of the promising encoder-decoder neural architectures of CO \citep{kool2018attention, kwon2020pomo, kim2022sym} at \cref{append:am}.

\paragraph{Guided exploration.} 
To encourage the policy assign more probabilities to candidates near the last action, we introduce a locality bias based on the distances.
Assume that the current state is $s_t = \{\boldsymbol{a}_{1:t};\boldsymbol{x}\}$, and the current location is $\bm{a}_t$ in TSP.
At each $t$, we can measure the distance between the remaining locations to visit with
\begin{equation*} 
    d(x_{\bm{a}_t},\bm{x}) := \left\{||x_i - x_{\bm{a}_t}||; i \in \{1,\ldots,N\} \setminus \bm{a}_{1:t} \right\}.
\end{equation*}
We induce the policy to select locations near the last visited location by penalizing the policy logit value $u_j$ for the next candidate $j$ as follows:
\begin{equation} \label{eq:bias}
    u_{j} \leftarrow u_{j} - \alpha d(x_{\bm{a}_t},\bm{x})_{j}, \mbox{ } \forall j \in \{1,\ldots,N\} \setminus \bm{a}_{1:t}.
\end{equation}

The motivation for locality bias is the lack of capability for global search in the early stages of adaptation. By inducing the model to search the local area first, the exploration capability is improved.

\subsection{Scheduled Adaptation} \label{sec:sage_op}

\paragraph{Test time adaptation.}

Following \citet{hottung2021efficient}, we update $\eta$ based on a single instance $\boldsymbol{x}$ with two objectives: reinforcement learning objective $\mathcal{J}_{RL}$ and self-imitation learning objective $\mathcal{J}_{IL}$.

$\mathcal{J}_{RL}$ is REINFORCE loss with shared baseline $b$ \citep{kwon2020pomo} which is evaluated with Monte Carlo sampling $\boldsymbol{a}_{(1)},\ldots,\boldsymbol{a}_{(M)} \sim p_{\theta^{*}}(\cdot|\boldsymbol{x}; \phi^*, \eta)$ given $\theta^*$ and $\phi^*$:
\begin{equation*}
\begin{split}
    &\mathcal{J}_{RL}(\eta;\bm{x}) = \\
    & \quad \frac{1}{M}\sum_{m=1}^{M} \left(R(\boldsymbol{a}_{(m)},\boldsymbol{x})-b \right) \cdot \log p_{\theta^{*}}(\boldsymbol{a}_{(m)}|\boldsymbol{x};\phi^*,\eta), \\
    &\quad \text{where} \quad b = \frac{1}{M}\sum_{m=1}^{M}R(\boldsymbol{a}_{(m)};\boldsymbol{x}).
\end{split}
\end{equation*}

$\mathcal{J}_{IL}$ is self-imitation loss to maximize the log likelihood of selecting the best-sampled solution $\boldsymbol{a}_{(*)} = \argmax \left\{R(\boldsymbol{a}_{(1)};\boldsymbol{x}),\ldots,R(\boldsymbol{a}_{(M)};\boldsymbol{x})\right\}$ as follows:
\begin{equation*}
    \mathcal{J}_{IL}(\eta;\bm{x}) = \log p_{\theta^{*}}(\boldsymbol{a}_{(*)}|\boldsymbol{x};\phi^*,\eta).
\end{equation*}
The total objective is then defined as:
\begin{equation} \label{eq:j_sage}
    \mathcal{J}_{\text{\adpt{}}}(\eta;\bm{x}) = \mathcal{J}_{RL}(\eta;\bm{x}) + \lambda\mathcal{J}_{IL}(\eta;\bm{x}), 
\end{equation}
where $\lambda$ is a tunable hyperparameter. The parameter $\eta$ is iteratively updated to maximize the total objective, i.e.,
\begin{equation*}
    \eta^{k} \leftarrow \eta^{k-1} + \delta \nabla \mathcal{J}_{\text{SAGE}}, \quad k=1, \ldots, K
\end{equation*} 
where $k$ is the adaptation iteration, and $\delta$ is a learning rate.

\paragraph{Scheduling.}
\adpt{} is an enhanced EAS with locality bias and scheduling. 
The locality bias coefficient $\alpha$ in \cref{eq:bias} is monotonically decreased as the model is adapted to the target instance (i.e., the adaptation step $k$ increases) as follows:
\begin{equation}
    \alpha \leftarrow \gamma_1 \alpha, \mbox{ where } 0 \leq \gamma_1 < 1.
\end{equation}
It gradually reduces the reliance on locality bias, a strong inductive bias for the local view. Also, the SoftMax temperature $\mathcal{T}$ in \cref{eq:policy_softmax} is scheduled for confidence exploitation as follows:

\begin{equation}
    \mathcal{T} \leftarrow \gamma_2 \mathcal{T}, \mbox{ where } 0 \leq \gamma_2 < 1.
\end{equation}

In summary, we get updated parameter via \adpt{} operation as illustrated in \cref{fig:structure},
i.e., $\eta^{K} \leftarrow \text{\adpt{}}(\bm{x}, K; \theta^{*},\phi^*)$ for the given parameters $\theta^{*},\phi^*$, and the target instance $\bm{x}$.
\section{Training \ours{}} \label{sec:learning}
This section provides a formulation of \ours{} and a meta-learning scheme for SML. We provide a pseudo-code in \cref{append:training_algo}. 

\subsection{Bi-level Formulation of Training \ours{}} \label{sec:SML}

The role of \meta{} is to give properly transformed embedding to the subsequent parameter adaptation (i.e., \adpt{}) based on the scale information.
We formulate our training of \ours{} as bi-level optimization as follows:
\begin{align}
    \label{eq:upper_opt}
    \max_{\textcolor{darkgreen}{\phi}} \quad  &  \mathbb{E}_{P(\bm{x})} \mathbb{E}_{p_{\theta^*}(\cdot|\bm{x};\phi, \textcolor{darkblue}{\eta})} \left[ R (\bm{a}_{1:N}, \bm{x}) \right]  \\
    \label{eq:lower_opt}
    \mbox{s.t.} \quad  & \textcolor{darkblue}{\eta} \leftarrow \text{\adpt{}}(\bm{x}, K; \theta^{*}, \textcolor{darkgreen}{\phi})
\end{align}

The upper-level optimization (i.e., training SML) in \cref{eq:upper_opt} contains $\eta$ that can be obtained by solving the lower-level optimization (i.e., SAGE). At the same time, the lower level requires optimized $\phi$ as an input. Therefore, the optimization problems are interrelated, making training more difficult.

\subsection{Meta learning with Distillation Scheme}
We propose Meta-learning with Distillation Scheme that reliably trains $\phi$, while breaking the inter-relationship between $\phi$ and $\eta$. 
The roles of $g_{\phi}(\bm{h}_{(l)},N)$ can be twofold: 
\begin{enumerate}
    \item Transporting the context embedding $\bm{h}_{(l)}$ to as close as possible to the instance-wise target embedding $\bm{h}^A_{(l)}=g_{\eta^K_{(l)}}(\bm{h}^S_{(l)})$. 
    \item Improving instance-wise adaptation quality with the scale-conditioned initial embedding $\bm{h}^{S}_{(l)} = g_{\phi}(\bm{h}_{(l)}, N)$.
\end{enumerate}

To achieve these two roles, we train SML parameter $\phi$ using the following objective:
\begin{equation} \label{eq:j_sml}
    \mathcal{J}_{\text{SML}}(\phi) = \mathcal{J}_{\text{distil}}(\phi) + \beta \mathcal{J}_{\text{zero}}(\phi),
\end{equation}
where $\beta$ is a tunable hyperparameter. The intuition behind each objective follows.

\textbf{Distillation objective.}
Let's assume that instance-wise target node embedding $\bm{h}^A_{(l)}$ can be obtained from the contextual embedding $\bm{h}_{(l)}$ regardless of the scale conditioned node embedding
if we conduct a sufficiently large number $K$ of SAGE adaptation on $\eta$ without SML parameters $\phi^*$, i.e., $\bm{h}^A_{(l)}=g_{\eta^{K}_{(l)}}(\boldsymbol{h}_{(l)})$, where $\eta^{K}_{(l)} \leftarrow \text{\adpt{}}(\bm{x}, K; \theta^{*})$. Then, given $\bm{x}_{(l)} \sim P(\bm{x}|N)P(N)$ for $l=1,\ldots,L$, we can amortize $g_{\phi}(\bm{h}_{(l)},N)$ to imitate $g_{\eta^{K}_{(l)}}(\bm{h}_{(l)})$, which is referred to as distillation. This can be done by maximizing the following objective:
\begin{equation}
    \mathcal{J}_{\text{distil}} (\phi) = -\frac{1}{L}\sum_{l=1}^{L}\|g_{\phi}(\boldsymbol{h}_{(l)},N_{(l)}) - g_{\eta^{K}_{(l)}}(\boldsymbol{h}_{(l)} )\|_2
\end{equation}
with sufficiently large $K$. Note that by maximizing $\mathcal{J}_{\text{distil}} (\phi)$, the distance between the scale-conditioned embedding $\bm{h}^S_{(l)}$ and the instance-wise target embedding decreases so that
\begin{equation*}
    \bm{h}^S_{(l)} = g_\phi(\bm{h}_{(l)}, N_{(l)}) \approx \mathbb{E}_{P(\bm{x}|N=N_{(l)})} g_{\eta^K_{(l)}} (\bm{h}_{(l)}).
\end{equation*}
That is, SML is trained to convert context embedding  to the expected target embedding $\bm{h}_{(l)}^T=g_{\eta^K_{(l)}} (\bm{h}_{(l)})$ for the problem instances $\bm{x}_{(l)} \sim P(\bm{x}|N)P(N)$ for $l=1,\ldots,L$.

\textbf{Zero-shot simulation objective. }
We hope that the scale-conditioned initial embedding $\bm{h}^S=g_{\phi}(\bm{h}, N)$ to be a sufficiently good representation to solve a target problem instance $\bm{x}$ even before $\bm{h}^S$ is transformed to the instance-wise target embedding $\bm{h}^A$. Thus, it can be interpreted as the input $\bm{h}^S$ is the $0$-th adapted target embedding (i.e., SAGE with $K = 0$).
Based on this interpretation, we define the zero-adaptation objective as follows:
\begin{equation*}
    \mathcal{J}_{\text{zero}}(\phi) = \mathbb{E}_{\bm{x} \sim P(\bm{x}|N)P(N)}\left[\mathcal{J}_{\text{SAGE}}(\phi;\bm{x}) \right].
\end{equation*}
The SAGE objective function $\mathcal{J}_{\text{SAGE}}$ is contextualized by $\bm{x}$. By taking an expectation over the problem distribution, the shared parameter $\phi$ is trained over the problem distribution $P(\bm{x}|N)P(N)$.

\begin{table*}[!h]
\centering
\caption{Performance evaluation results for the four CO tasks of TSP, CVRP, PCTSP, and OP are presented. We used two pre-trained DRL models, POMO and Sym-NCO (termed Sym. in the table), and deployed adaptation methods on each. The best-performing method is highlighted in bold among adaptation methods and marked as $*$ if the method outperforms the problem-specific solver. Note that OP is a maximization problem (i.e., the objective value is higher ($\uparrow$), the better), while others are minimization problems (i.e., the objective value is lower ($\downarrow$), the better). The reported time is for solving whole instances. } \label{Table01}
\scalebox{0.9}{\begin{tabular}{llllllllllll}
\toprule[1.2pt]
\multicolumn{2}{l}{\begin{tabular}{c}\multirow{2}{*}{}\end{tabular}}&&\multicolumn{3}{c}{$N = 200$ ($1$K Instances)}&\multicolumn{3}{c}{$N = 500$ ($128$ Instances)}&\multicolumn{3}{c}{$N = 1,000$ ($128$ Instances)}\\\cmidrule[0.5pt](lr{0.2em}){4-6} \cmidrule[0.5pt](lr{0.2em}){7-9} \cmidrule[0.5pt](lr{0.2em}){10-12}
&\multicolumn{2}{c}{}&\multicolumn{1}{l}{Obj.}&\multicolumn{1}{l}{Gap}&\multicolumn{1}{l}{ Time}&\multicolumn{1}{l}{Obj.}&\multicolumn{1}{l}{Gap}&\multicolumn{1}{l}{Time}
&\multicolumn{1}{l}{Obj.}&\multicolumn{1}{l}{Gap}&\multicolumn{1}{l}{Time}\\
\midrule[1.2pt]

 \parbox[t]{2mm}{\multirow{11}{*}{\rotatebox[origin=c]{90}{TSP ($\downarrow$)}}}

& \multicolumn{2}{c}{Concorde} & {10.687} & {0.000\%} & {0.4H}& {16.542} & {0.000\%}& {0.6H} &{23.139}&{0.000\%}&{5.4H}\\
\cmidrule(lr{0.1em}){2-11}
 \cmidrule(lr{0.1em}){2-12}
 & \parbox[t]{2mm}{\multirow{4}{*}{\rotatebox[origin=c]{90}{POMO}}}
&AS&10.735&0.449\%&22.4H&17.335&4.791\%&9.2H&-&-&-\\
&&EAS  &10.736&0.455\%&2.4H&18.135&9.362\%&4.3H&30.744& 32.869\%&20H\\
&&SBGS  &10.734&0.436\%&2.1H&18.191&9.963\%&4.2H&28.413&22.795\%&19H\\
\cmidrule(lr{0.1em}){3-12}
&&{Ours}  &\textbf{10.729}&\textbf{0.391\%}&\textbf{2.1H}&\textbf{17.131}&\textbf{3.559\%}&\textbf{3.8H}&\textbf{25.924}&\textbf{12.038\%}&\textbf{18H}\\
 \cmidrule(lr{0.1em}){2-12}
  & \parbox[t]{2mm}{\multirow{4}{*}{\rotatebox[origin=c]{90}{Sym.}}}
&AS &10.748&0.575\%&22.4H&17.352&4.897\%&9.2H&-&-&-\\
&&EAS&10.731&0.413\%&2.4H&18.194&9.986\%&4.3H&31.241&35.017\%&20H\\
&&SBGS  &10.730&0.402\%&2.1H&18.193&9.98\%&4.2H&28.431& 22.87\%&19H\\
\cmidrule(lr{0.1em}){3-12}
&&{Ours}  &\textbf{10.728}&\textbf{0.387\%}&\textbf{2.1H}&\textbf{17.095}&\textbf{3.339\%}&\textbf{3.8H}&\textbf{25.798}&\textbf{11.488\%}&\textbf{18H}\\

\midrule[1.2pt]

 \parbox[t]{2mm}{\multirow{11}{*}{\rotatebox[origin=c]{90}{CVRP ($\downarrow$)}}}
& \multicolumn{2}{c}{LKH3}  &22.003&0.000\%&25H&{63.299}&{0.000\%}&16H&{120.292}&{0.000\%}&40H\\
 \cmidrule(lr{0.1em}){2-12}
  & \parbox[t]{2mm}{\multirow{4}{*}{\rotatebox[origin=c]{90}{POMO}}}
&AS &22.050&0.213\%&28.2H&64.053&1.192\%&\text{11.5H}&-&-&- \\

&&EAS  &22.023&0.091\%&3.1H&64.318&1.610\%&\text{5.5H}&125.043&3.95\%&\text{24H}\\
&&SBGS  &22.015&0.055\%&3.1H&65.211&3.02\%&\text{5.2H}&126.554&5.206\%&20H\\
\cmidrule(lr{0.1em}){3-12}
&& {Ours} &\textbf{\text{22.001}*}&\textbf{\text{-0.009\%}*}&\textbf{2.7H}*&\textbf{63.322}&\textbf{0.035\%}&\textbf{4.7H}&\textbf{121.114}&\textbf{0.683\%}&\textbf{20H}\\
 \cmidrule(lr{0.1em}){2-12}
  & \parbox[t]{2mm}{\multirow{4}{*}{\rotatebox[origin=c]{90}{Sym.}}}
&AS &$\textbf{\text{21.955}*}$&\textbf{\text{-0.218\%}*}&28.2H&63.308&0.015\%&11.5H&-&-&- \\

&&EAS  &22.038&0.159\%&3.1H&64.256&1.511\%&5.5H&124.711&3.674\%&24H\\
&&SBGS  & 22.029 &0.116\%&  3.1H  &65.163&2.945\%&5.2H&126.208&4.918\%&20H\\
\cmidrule(lr{0.1em}){3-12}
&&{Ours}  &\text{21.982}&\text{-0.094\%}&\textbf{2.7H}*&\textbf{\text{63.281}*}&\textbf{\text{-0.028\%}*}&\text{\textbf{4.7H}}*&\textbf{121.116}&\textbf{0.685\%}&\textbf{20H}\\

\midrule[1.2pt]

 \parbox[t]{2mm}{\multirow{4}{*}{\rotatebox[origin=c]{90}{PCTSP ($\downarrow$)}}}
 
& \multicolumn{2}{c}{OR-Tools}  &7.954&0.000\%&16.7H&12.201&0.000\%&2.1H&23.611&0.000\%&2.1H\\
\cmidrule(lr{0.1em}){2-12}
  & \parbox[t]{2mm}{\multirow{3}{*}{\rotatebox[origin=c]{90}{Sym.}}}
  &AS &7.325&-7.8\%&3.5H&11.589&-5.013\%&2H&17.172&-27.273\%&3.9H \\
&&EAS  &\text{7.335}*&\text{-7.781\%}*&\text{9M}*&12.369&1.378\%&5.4M&\text{21.076}*&\text{-10.737\%}*&0.3H*\\
\cmidrule(lr{0.1em}){3-12}
&&{Ours}  &\textbf{\text{7.299}*}&\textbf{\text{-8.236\%}*}&\textbf{\text{6.6M}*}&\textbf{\text{11.175}*}&\textbf{\text{-8.412\%}*}&\textbf{3.7M}*&\textbf{\text{16.219}*}&\textbf{\text{-31.31\%}*}&\textbf{0.2H}*\\

\midrule[1.2pt]

 \parbox[t]{2mm}{\multirow{4}{*}{\rotatebox[origin=c]{90}{OP ($\uparrow$)}}} 
  & \multicolumn{2}{c}{Compass} &{49.216}&{0.000\%}&20.9M&97.323&0.000\%&11M&{161.682}&{0.000\%}&48.8M\\
\cmidrule(lr{0.1em}){2-12}
  & \parbox[t]{2mm}{\multirow{3}{*}{\rotatebox[origin=c]{90}{Sym.}}}
  &AS &48.946&0.548\%&2.8H&97.696&-0.382\%&1.1H&154.233&4.607\%&1.7H \\
&&EAS  &48.849&0.746\%&7.2M&85.032&12.63\%&2.8M&107.75&33.357\%&7.4M\\
\cmidrule(lr{0.1em}){3-12}
&&{Ours}  &\textbf{49.05}&\textbf{0.337\%}&\textbf{5.4M}&\textbf{98.762}*&\textbf{-1.457\%}*&\textbf{2.2M}*&\textbf{154.307}&\textbf{4.562\%}&\textbf{6.5M}\\
\bottomrule[1.2pt]
\end{tabular}}
\end{table*}

\section{Experimental Results}

This section provides experimental results to validate \ours{} with two pre-trained DRL models on four CO tasks. See \cref{append:imp_detail_ours} for the details of implementation.

\textbf{DRL models for adaptation.} Our new adaptation method is deployed to two representative DRL models for CO: Policy Optimization for Multiple Optima \citep[POMO; ][]{kwon2020pomo} and Symmetric Neural Combinatorial Optimization \citep[Sym-NCO; ][]{kim2022sym}. We use the pre-trained models on $N=100$ published online\footnote{Pre-trained models are available at \url{https://github.com/yd-kwon/SGBS} and \url{https://github.com/alstn12088/Sym-NCO}}; thus, the performance of these two models is supposed to be not good on larger-scale instances without adaptation.  

\textbf{CO Tasks.} We target four CO tasks: the traveling salesman problem (TSP), the capacitated vehicle routing problem (CVRP), the Prize Collecting TSP (PCTSP), and the orienteering problem (OP). The tasks are known as NP-hard \citep{garey1979computers}.
\begin{itemize}
    \item The TSP is to find the shortest route for a salesman to visit every city and return to the first city, also known as a Hamiltonian cycle with minimum distances.  
    \item The CVRP \citep{dantzig1959truck} is a variation of the TSP that assumes multi-vehicles with limited capacity. CVRP aims to find the set of tours each of which starts from the depot, visits cities once, and returns to the depot and the union of the tours covers all cities while minimizing the total distances.
    
    \item In PCTSP \citep{balas1989prize}, 
    each city has a prize and penalty; thus, the salesman gets prizes for visiting the cities and penalties for the unvisited cities. PCTSP minimizes the total length of the route and the net penalties while collecting at least the minimum prizes.
    
    \item Lastly, the OP \citep{golden1987orienteering} also considers prizes associated with visiting. The goal is to find a route that maximizes the prizes from visited cities while keeping the length of the route shorter than a given maximum distance.
\end{itemize}

POMO and Sym-NCO models are pre-trained on TSP and CVRP, and Sym-NCO is additionally pre-trained on PCTSP and OP. Note that these TSP variant problems are highly challenging. 

\begin{figure*}
     \centering
     \begin{subfigure}[b]{0.32\textwidth}
         \centering
         \includegraphics[width=\textwidth]{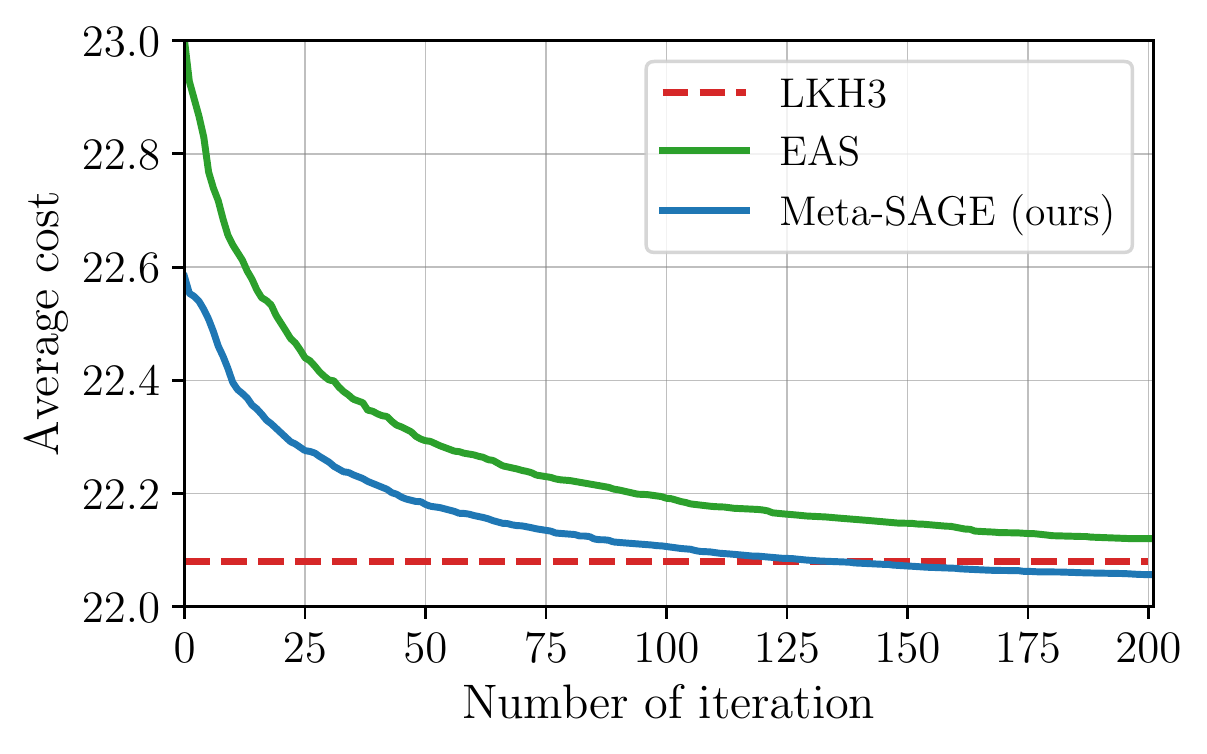}
         \caption{$N=200$}
         \label{fig:perf_shot_cvrp_200}
     \end{subfigure}
     \begin{subfigure}[b]{0.32\textwidth}
         \centering
        \includegraphics[width=\textwidth]{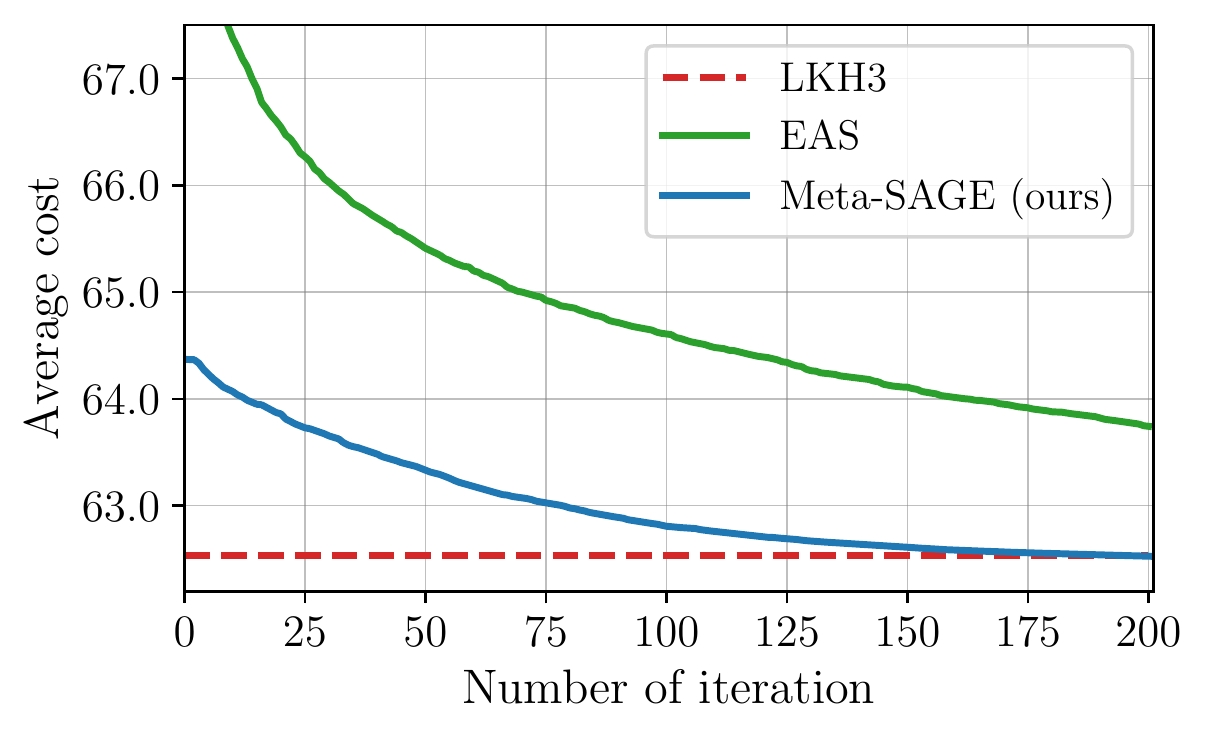}
         \caption{$N=500$}
         \label{fig:perf_shot_cvrp_500}
     \end{subfigure}
          \begin{subfigure}[b]{0.32\textwidth}
         \centering
        \includegraphics[width=\textwidth]{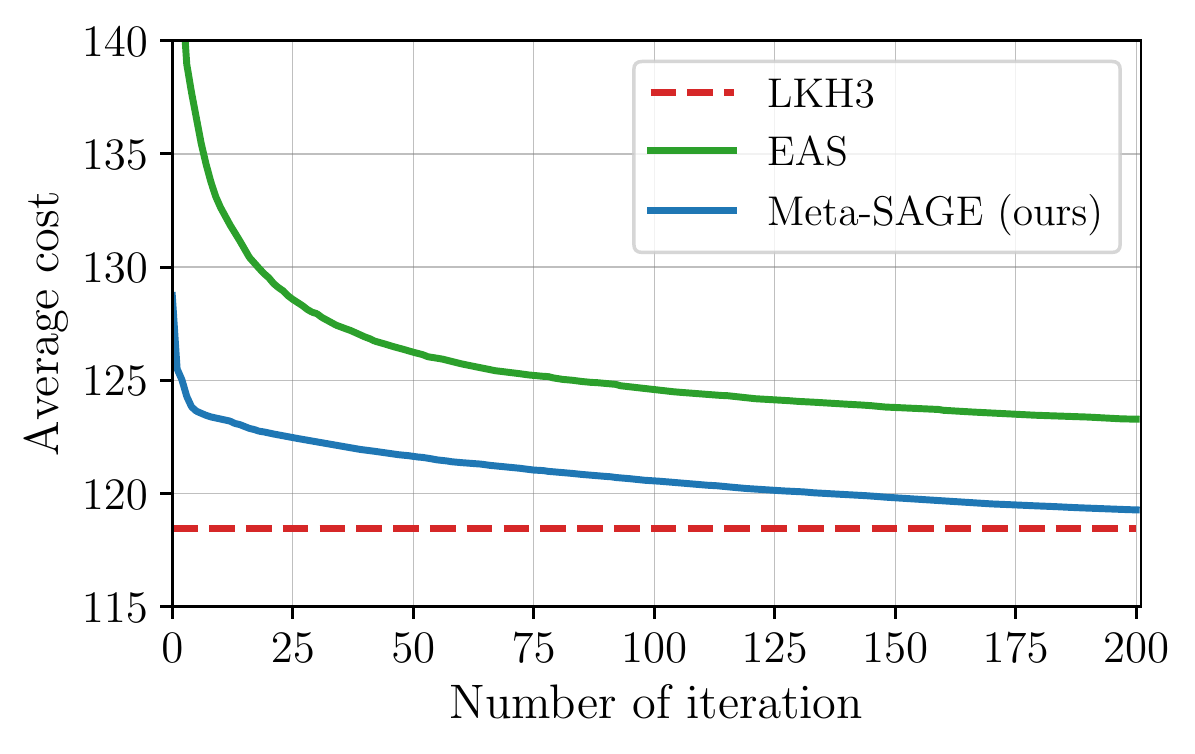}
         \caption{$N=1,000$}
         \label{fig:perf_shot_cvrp_1000}
     \end{subfigure}
     \caption{Adaptation performance on CVRP with pre-trained Sym-NCO compared to LKH3.}
    \label{fig:perf_shot_cvrp}
\end{figure*}
\begin{figure*}[t!]
     \centering
     \begin{subfigure}[b]{0.32\textwidth}
         \centering
         \includegraphics[width=\textwidth]{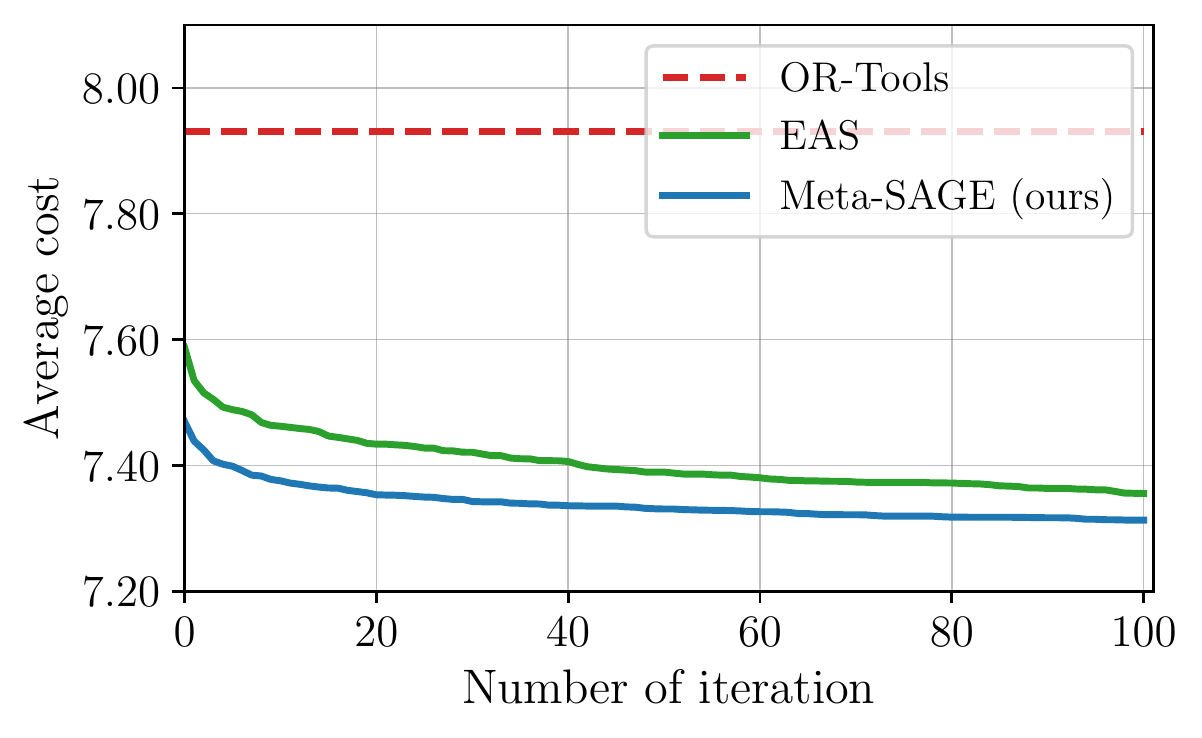}
         \caption{$N=200$}
         \label{fig:perf_shot_pctsp_200}
     \end{subfigure}
     \begin{subfigure}[b]{0.32\textwidth}
         \centering
        \includegraphics[width=\textwidth]{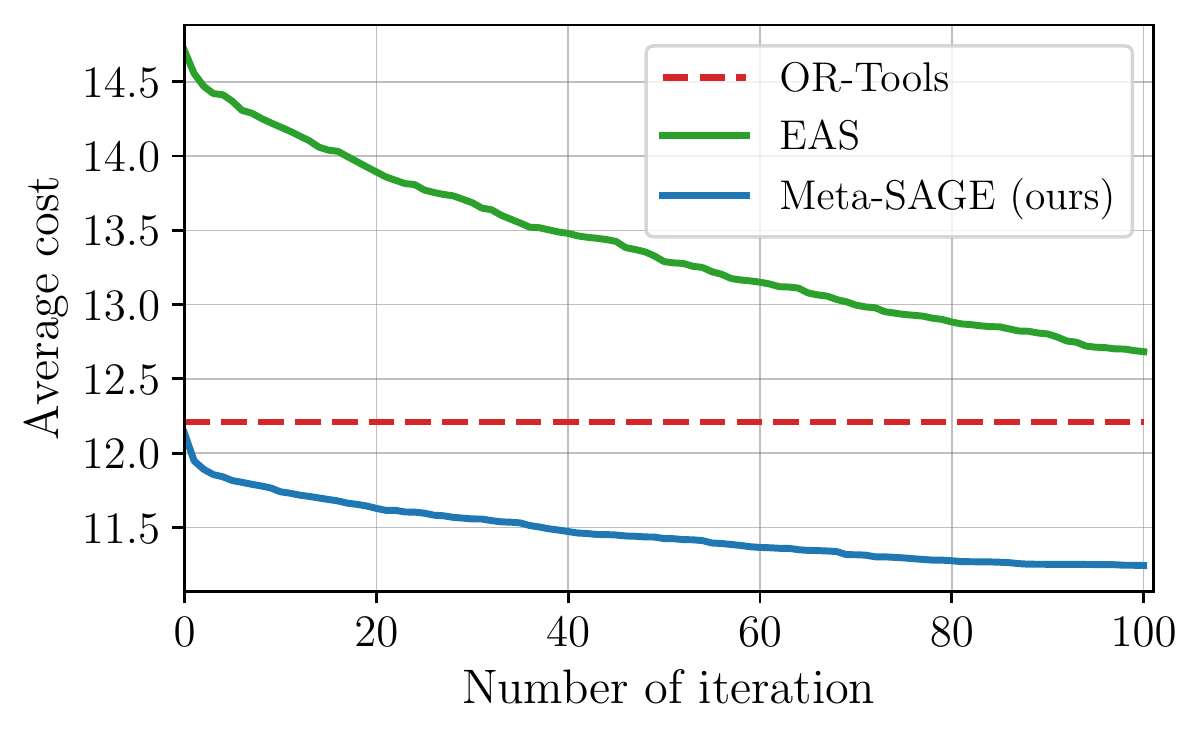}
         \caption{$N=200$}
         \label{fig:perf_shot_pctsp_500}
     \end{subfigure}
          \begin{subfigure}[b]{0.32\textwidth}
         \centering
        \includegraphics[width=\textwidth]{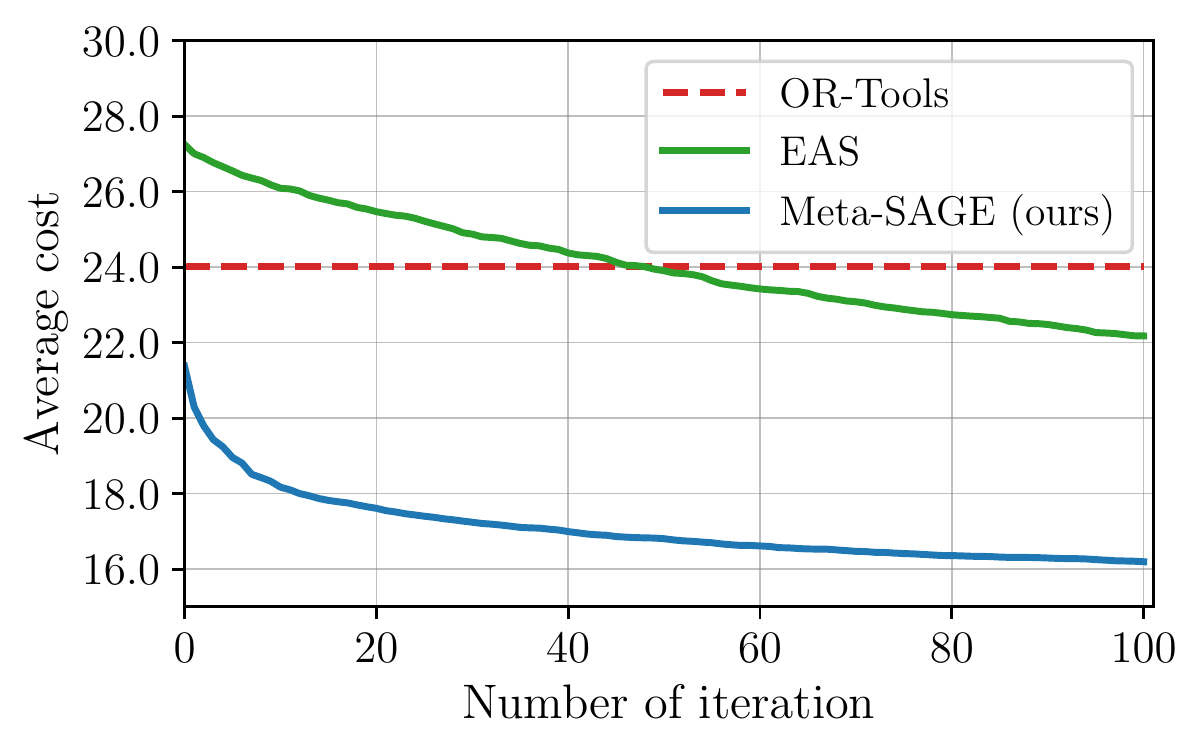}
         \caption{$N=1,000$}
         \label{fig:perf_shot_pctsp_1000}
     \end{subfigure}
     \caption{Adaptation performance on PCTSP with pre-trained Sym-NCO compared to OR-Tools.}
    \label{fig:perf_shot_pctsp}
\end{figure*}

\textbf{Baselines.} We consider three baselines for test time adaptation: active search \citep[AS; ][]{bello2017neural}, Efficient Active Search \citep[EAS; ][]{hottung2021efficient} and Simulation Guided Beam Search with EAS \citep[SGBS; ][]{choo2022simulation}. We describe the details of implementation in \cref{append:imp_detail_baseline}.

\textbf{Performance metric. } We measure the Baseline Gap using state-of-the-art solvers for each task.
As baseline solvers, we employ Concorde \citep{concorde} for TSP, LKH3 \citep{lkh2017} for CVRP, OR-Tools \citep{ortools} for PCTSP, and Compass \citep{kool2018attention} for OP\footnote{While Hybrid Genetic Search \citep[HGS;][]{vidal2022hybrid} is considered SOTA for CVRP, we employ LKH3 following the convention. For PCTSP, there is an Iterative Local Search \citep[ILS;][]{ils}, but OR-Tools performs better at large scale within reasonable time budgets \citep{kim2021learning}.}.

\textbf{Computation resources and runtime. } We used a single NVIDIA A100 40GB VRAM GPU and an Intel Scalable Gold CPU for the experiments. 
We terminate EAS and SGBS when the runtime exceeds the runtime of \ours{} with $200$ iterations.
Since AS is unavailable to process in batch, we conduct 200 iterations for adaptation, the same as ours for each instance.
We measure the time taken to solve whole instances following \citet{hottung2021efficient}.

\subsection{Main Results}

As shown in \cref{Table01}, \ours{} outperforms other adaptation baselines except for Sym-NCO on CVRP with 200. The results show that our method improves the pre-trained DRL models with EAS in consistency. 
Surprisingly, \ours{} finds better solutions than the heuristic solver on CVRP with $N=200, 500$, PCTSP with every scale, and OP with $N=500$. 
Note that heuristic solvers are designed for each problem class and are highly engineered. 

It is noteworthy that \ours{} transforms the pre-trained model to outperform the task-specific heuristic solver.

\cref{fig:perf_shot_cvrp} and \cref{fig:perf_shot_pctsp} illustrate the cost obtained by EAS and \ours{} with respect to the adaptation iteration on CVRP and PCTSP. 
The results clearly show that our method achieves improved performances compared to EAS with the same number of iteration (i.e., improved transferability). For example, on CVRP with $N=200, 500$, \ours{} outperforms LKH3 with 200 iterations, whereas EAS still gives higher costs than LKH3.
Noticeably, our initial (i.e., zero-shot) cost is lower than EAS in every instance, meaning that SML gives the scale-conditioned initial embedding $\bm{h}^S$ by appropriately transforming the original context embedding $\bm{h}$ before parameter adaptation. See \cref{append:time-performance} for the results of TSP and OP.

\subsection{Real-world Benchmark}

\begin{table}
\centering
\caption{Evaluation of adaptation methods with pre-trained Sym-NCO on the real-world CVRP dataset. We report average values according to the range of instance scale $N$ between $500$ and $1,000$. The full results are provided in \cref{append:cvrplib_full}.} \label{cvrplib_rst}
\scalebox{0.9}{\begin{tabular}{cccc}
\toprule[1.0pt]
\multirowcell{2}[-0.7ex]{Range} &\multirowcell{2}[-0.7ex]{LKH3} & \multicolumn{2}{c}{Sym-NCO}
\\
\cmidrule[0.5pt](lr{0.2em}){3-4}
 & & EAS & Ours 
\\
\specialrule{0.5pt}{2pt}{4pt}

$500 \leq N < 600$ &92,095&95,384& \textbf{94,028}\\
$600 \leq N < 700$ &85,453&88,987& \textbf{87,372}\\
$700 \leq N < 800$ &89,009&92,288& \textbf{90,184}\\
$800 \leq N < 900$ &112,785&119,332& \textbf{116,096}\\
$900 \leq N \leq 1000$&150,332&175,999& \textbf{172,916}\\

\bottomrule[1.0pt]
\end{tabular}}
\vspace{-15pt}
\end{table}

\begin{figure*}
     \centering
     \begin{subfigure}[b]{0.32\textwidth}
         \centering
         \includegraphics[width=\textwidth]{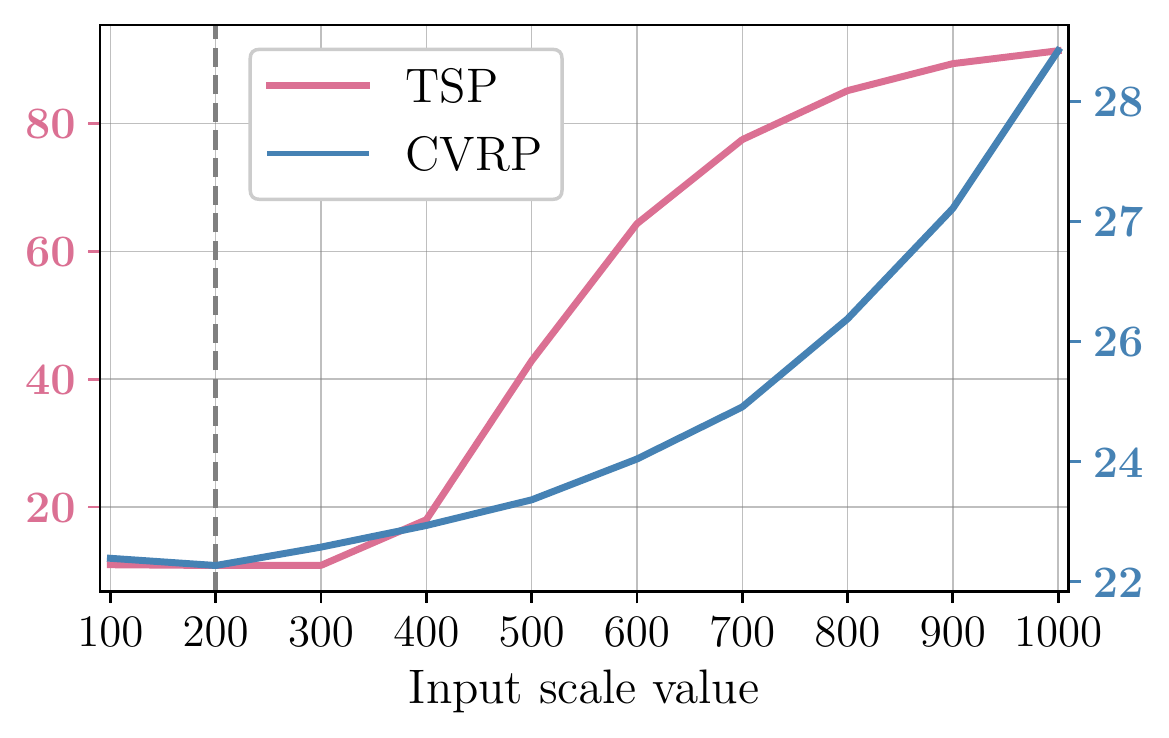}
         \caption{$N=200$}
         \label{fig:scale_200}
     \end{subfigure}
     \begin{subfigure}[b]{0.32\textwidth}
         \centering
        \includegraphics[width=\textwidth]{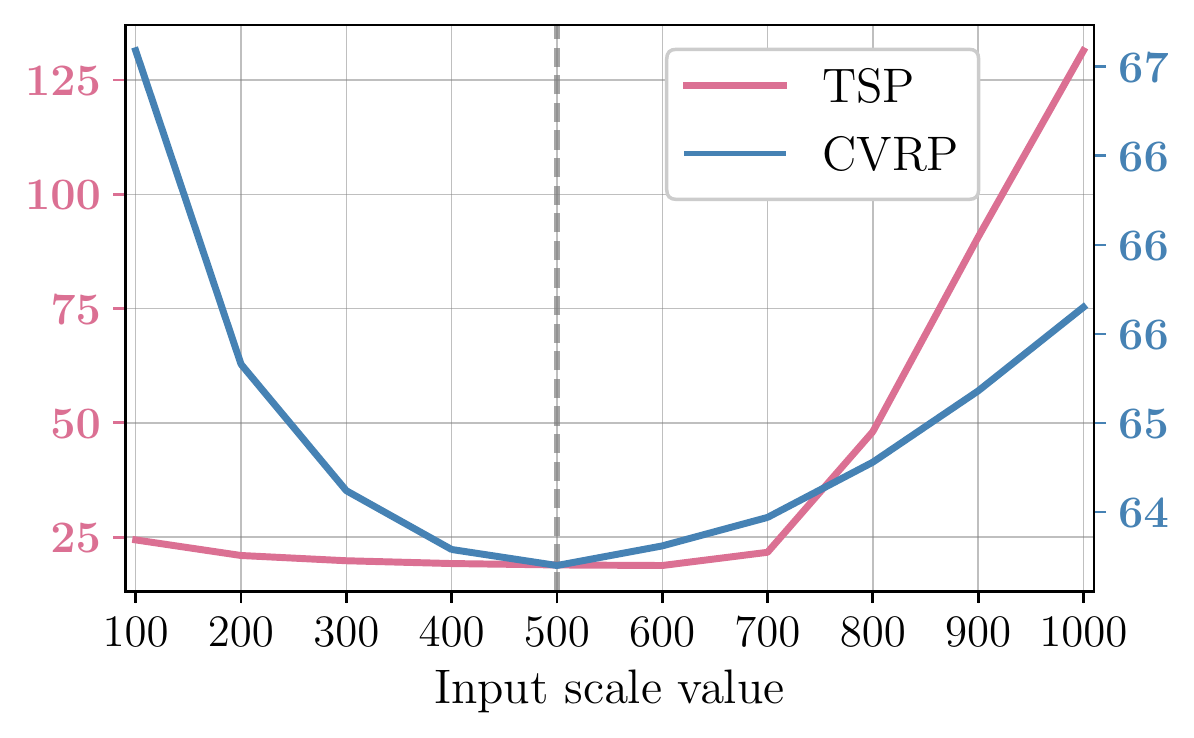}
         \caption{$N=500$}
         \label{fig:scale_500}
     \end{subfigure}
          \begin{subfigure}[b]{0.32\textwidth}
         \centering
        \includegraphics[width=\textwidth]{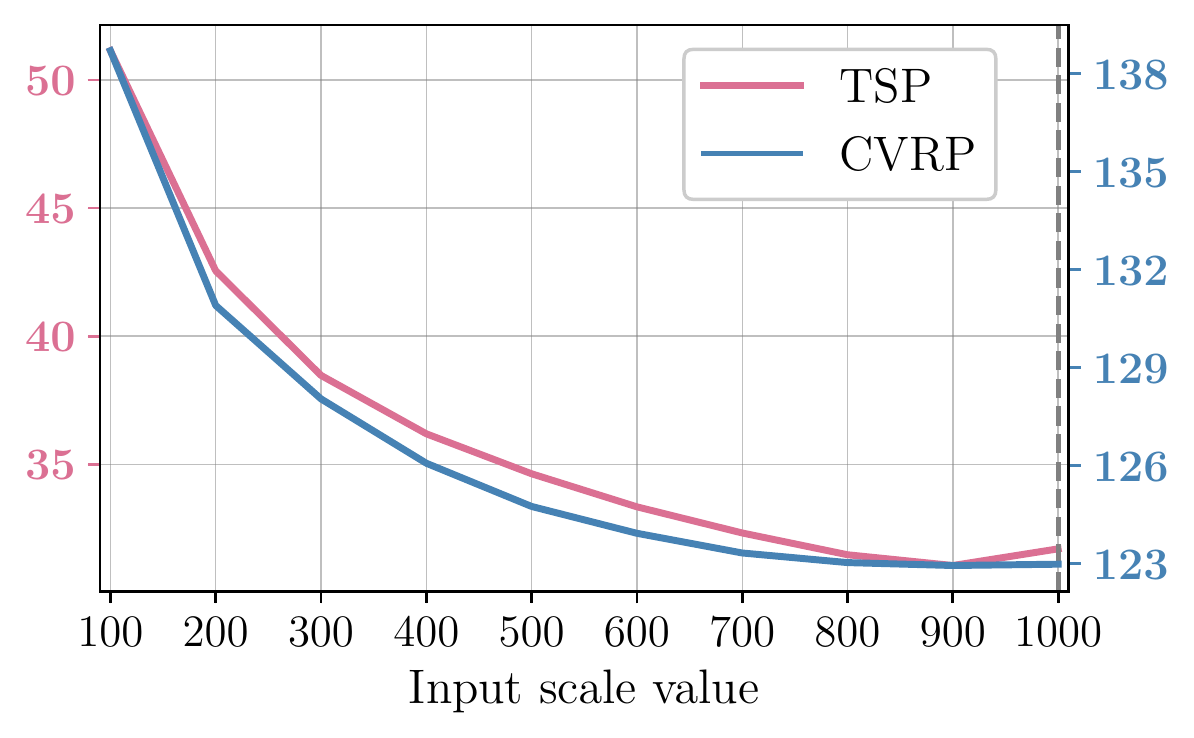}
         \caption{$N=1,000$}
         \label{fig:scale_1000}
     \end{subfigure}
     \vspace{-5pt}
     \caption{Scale matching performance of SML. For each target scale of $N=200, 500, 1000$, SML gives lower cost (i.e., better performances) when the input scale $N'$ is closer to the actual target scale.}
    \label{figure:sml-calibration}
    \vspace{-12pt}
\end{figure*}

We apply our method on the real-world dataset for CVRP, X-instances in CVRPLIB \citep{uchoa2017new}. The model is trained with CVRP instances whose demands and locations are samples from uniform distributions. On the other hand, the X-instances are generated by various demand and location distributions to cover real-world applications. 
As shown in \cref{cvrplib_rst},
Meta-SAGE outperforms EAS in every range. Moreover, the real-world experiment demonstrates that our method can overcome the distribution shift.

\subsection{Ablation Study}

\begin{table}
\centering
\caption{{Ablation study for \ours{} component. LB and Temp. refer to locality bias and SoftMax temperature, respectively.}}\label{abl:sml-sage}
\scalebox{0.9}{
\begin{tabular}{cccccc}
\toprule[1.0pt]
\multirowcell{2}[-0.7ex]{SML} & \multicolumn{3}{c}{SAGE} &  \multicolumn{2}{c}{Tasks}\\
\cmidrule(lr{0.2em}){2-4}\cmidrule(lr{0.2em}){5-6}
 & LB & Sche. LB & Sche. Temp. & TSP & CVRP \\

\midrule
 & & & &18.63&63.97 \\
\checkmark  & & & &17.70&63.17 \\
\checkmark  & \checkmark & & &17.68&63.10 \\
\checkmark  & \checkmark & \checkmark & &{17.67}&{63.07} \\
\checkmark  & \checkmark & \checkmark  & \checkmark &\textbf{17.22}&\textbf{62.69} \\

\bottomrule[1.0pt]
\end{tabular}
\vspace{-10pt}
}
\end{table}

\textbf{Effectiveness of main components. } \ours{} consists of SML and SAGE composed of guided exploration with locality bias and scheduling. The SAGE has two scheduling targets, locality bias and SoftMax temperature. The performances are measured while ablating each element. The experiments are conducted on TSP and CVRP with $N=500$, using the Sym-NCO as a pre-trained DRL model. Each component significantly impacts performance improvement, as shown in the \cref{abl:sml-sage}.

\textbf{Effectiveness of SML. }
The objective of SML is to support SAGE by providing proper scale-conditioned initial embedding $\bm{h}^{S} = g_{\phi}(\bm{h}, N)$ based on the scale information. To verify SML's effect, we measure the $L1$ distance between the input and output embedding of SAGE. The effectiveness can be demonstrated by the reduced distance by SML, i.e.,

\begin{equation*}
||\bm{h}^{S} - \bm{h}^A|| < ||\bm{h} - \bm{h}^A||.
\end{equation*}

\begin{table}
\centering
\caption{The $L1$ distance between input and output embedding of SAGE. The distances are measured with and without SML using Sym-NCO.
}\label{abl:sim-measure}
\scalebox{0.9}{\begin{tabular}{ccc}
\toprule[1.0pt]
& TSP ($N=500$)& CVRP ($N=500$)\\
\midrule
w/ SML& \textbf{7,175.99} & \textbf{9,285.22}\\
w/o SML& 8,066.13 &  10,361.51\\

\bottomrule[1.0pt]
\end{tabular}
}
\vspace{-15pt}
\end{table}

As shown in \cref{abl:sim-measure}, SML reduces the distance between the initial and target embedding of SAGE, which verifies that SML provides effective initial embedding for SAGE.  

\textbf{Scale matching abilities of SML. } \ours{} improves adaptation efficiency by capturing scale information via SML. Therefore, the best performance of SML should be obtained when the scale input matches the target scale.
\cref{figure:sml-calibration} illustrates that SML performs well when the input scales are close to the actual target scale.

\textbf{Effectiveness of objective components in SML.} We verify the effectiveness of two objective function in \cref{eq:j_sml}; empirical results can be found in \cref{distil_abl}.

\section{Discussion}

\subsection{Other Meta-learning for CO}
\citet{qiu2022dimes} suggested Differentiable Meta Solver for Combinatorial Optimization Problems (DIMES) by utilizing the model agnostic meta-learning \citep[MAML; ][]{finn2017model} to compensate for instance-wise search. 

The difference is that we focus more on scale transferability, which is beneficial when the model is unavailable to be pre-trained on large scales. Our approach utilizes already trained (i.e., not meta-trained) models and adapts them to unseen tasks by considering scale prior. In addition, our method can be orthogonally applied with pre-trained DIMES for further improvements. 

\subsection{Extension to Non-Euclidean Graph CO}
We verified that Meta-SAGE achieves promising performances on large-scale Euclidean CO problems by adapting the pre-trained model. However, several important problems, such as scheduling, are non-Euclidean CO, whose edge values cannot be obtained with node values. Meta-SAGE can be extended by being attached to promising pre-trained models for non-Euclidean CO; we leave it as further work.

\subsection{Distributional Shifts}
This work focuses on the distributional shift of problem scale. However, the robustness for other distributional shifts, including instance distribution, is required to make DRL solvers more practical, especially when they are trained with fixed known distributions. Despite several prior research \citep{jiang2022learning,xin2022generative,bilearning}, tackling distributional shifts remains challenging. Our additional experiments on real-world CVRP benchmarks show the feasibility of Meta-SAGE on instance-wise distributional shifts. We expect that Meta-SAGE can consider other distributional shifts, not only scale. 

\subsection{Mathematical Programming}

While neural combinatorial optimization (NCO) approaches provide near-optimal solutions with fast computation, it cannot be claimed that they exhibit superiority over mathematical programming approaches such as branch-and-bound. Unlike NCO, mathematical programming-based approaches, such as branch-and-bound, are theoretically guaranteed to find optimal solutions. Since current NCO solvers cannot guarantee optimal solutions or provable optimality gaps, the theoretical study of NCO remains a challenging and significant area for future exploration.

\section*{Acknowledgement}

We thank Fanying Chen, Kyuree Ahn, and anonymous reviewers for providing helpful feedback for preparing our manuscripts. This work was supported by a grant of the KAIST-KT joint research project through AI2XL Laboratory, Institute of convergence Technology, funded by KT [Project No. G01210696, Development of Multi-Agent Reinforcement Learning Algorithm for Efficient Operation of Complex Distributed Systems].

\bibliography{reference.bib}
\bibliographystyle{utils/icml2023.bst}

\newpage
\onecolumn
\appendix

\section{Trainig Algorithm for \ours{}}
\label{append:training_algo}
\begin{algorithm}
\caption{Meta-SAGE}\label{alg:meta-sage}
\begin{algorithmic}
\STATE Set $\theta^{*} \leftarrow \text{pretrained model parameter}$, $\phi$: SML parameter. 
\\\hrulefill

\STATE \textit{A. Meta-Learning SML $\phi$}
\FOR{$t = 1.,,,.T$}
\STATE $\mathcal{J}_{\text{zero}} \leftarrow$ Zero-shot simulation objective of SAGE
\STATE $\mathcal{J}_{\text{distil}} \leftarrow$ Distil. Objective with pre-sim. of SAGE

\STATE Train $\phi \leftarrow \phi_{t}$ to maximize $\mathcal{J}_{distil} + \beta \mathcal{J}_{\text{zero}}$

\ENDFOR
\\\hrulefill
\STATE Set $\boldsymbol{x}_{\text{test}}$, where $N$ is large. \\Set $\eta$ is a test-time learnable parameter. \\Set $\mathcal{T}$: initial SoftMax temperature. \\Set $\alpha$: initial locality bias level.\\Set $0<\gamma_1<1$ and $0<\gamma_2<1$ are decaying factors.\\ 
Set $f_{\theta^{*}}(\bm{x})$ as the pre-trained encoder.\\
Set $p_{\theta^{*}}(\bm{a}|\bm{h};\mathcal{T}, \alpha)$ as the pre-trained decoder.\\
Set $g_{\phi^{T}}(\bm{h},N)$ as the trained SML.\\
\hrulefill
\STATE \textit{B. Test-time Adaptation with SAGE}
\FOR{$k = 1.,,,.K$}
\STATE $\bm{h} \leftarrow f_{\theta^{*}}(\bm{x};\theta^{*})$: Context embedding from Encoder.
\STATE $\bm{h} \leftarrow g_{\phi^{T}}(\bm{h},N)$: \textbf{SML's Context Transformation.}
\STATE $\bm{h} \leftarrow g_{\eta}(\bm{h})$: Test-time Context Transformation
\STATE $\boldsymbol{a}^1,...,\boldsymbol{a}^M \sim p_{\theta^{*}}(\cdot|\bm{h};\mathcal{T}, \alpha)$: MC sampling
\STATE $\mathcal{J}_{RL} \leftarrow$ REINFORCE objective with shared baseline
\STATE $\mathcal{J}_{IL} \leftarrow$ Self Imitation objective
\STATE Train $\eta \leftarrow \eta_{k}$ to maximize $\mathcal{J}_{RL} + \lambda \mathcal{J}_{IL}$
\STATE \textbf{Scheduling}: $\alpha \leftarrow \gamma_1 \alpha$, $\mathcal{T} \leftarrow \gamma_2 \mathcal{T}$, 

\ENDFOR

\end{algorithmic}
\end{algorithm}

\section{Implementation Details of \ours{}} \label{append:imp_detail_ours}

\subsection{Architecture of Target Model} \label{append:am}

In this research, we focus on two representative DRL models, POMO \citep{kwon2020pomo} and Sym-NCO \citep{kim2022sym}. Both models have a similar neural architecture, based on the attention model \citep[AM]{kool2018attention}. The main difference between these models is the training method for the AM.

The AM has a transformer-style encoder-decoder structure, where the encoder $f_{\text{enc}}(\bm{x})$ creates a context embedding $\bm{h}$ and the decoder $p_{\text{dec}}(\bm{h})$ generates action sequences $\bm{a}_{1:N}$ in an auto-regressive manner. The architecture of the encoder is not discussed here as our method and EAS \citep{hottung2021efficient} focus on $\bm{h}$ for the decoding scheme. For more information on the encoder architecture, refer to \citep{kool2018attention}.

The decoder $p_{\text{dec}}(\bm{h})$ is an auto-regressive process, which produces $\bm{a}t$ by referring to previously selected actions $\bm{a}_{1:t-1}$. Specifically, $\bm{h}$ is transformed into key $\bm{K}$ and value $\bm{V}$ via linear projection.

\begin{align*}
\bm{K} = \Phi_{\text{linear},k}(\bm{h}), \quad \bm{V}=\Phi_{\text{linear},v}(\bm{h})\text {. }
\end{align*}

Then the contextual query $q_c$ is computed as follows:

\begin{align*}
\bm{q}_c = \frac{1}{N}\sum_{i=1}^{N}{\bm{h}_i} + \bm{h}_{\text{depot}} + \bm{h}_{\bm{a}_{t-1}}\text {. }
\end{align*}

The $h_{\text{depot}}$ indicates the embedding of the depot node, and $\bm{h}_{\bm{a}_{t-1}}$ indicates the embedding of the previously selected action. 
Then we can compute compatibility $u_j$ as:


\begin{equation}
    u_j = \begin{cases}
    C\tanh\left(\frac{q_{c}^{T}K_j}{\sqrt{d_k}}\right) & \mbox{ if } j \notin \bm{a}_{1:t-1} \\
    - \infty & \mbox{ otherwise.}
    \end{cases}\label{eq:compat}
\end{equation}

Using compatibility, we can compute attention weight as:

\begin{equation*}
a_{j}=\frac{e^{u_{j}}}{\sum_{j^{\prime}} e^{u_{j^{\prime}}}} \text {. }
\end{equation*}

Then finally, the contextual query is updated as:

\begin{equation}
\mathbf{q}^{*}_c =  \sum_j a_{j} \mathbf{v}_j\text {. }
\end{equation}

Then finally, the compatibility for selecting the next action is computed the same with \cref{eq:compat} but using $\mathbf{q}^{*}_c$:

\begin{equation}
    u^{*}_j = \begin{cases}
    C\tanh\left(\frac{\mathbf{q}^{*^T}_{c}K_j}{\sqrt{d_k}}\right) & \mbox{ if } j \notin \bm{a}_{1:t-1} \\
    - \infty & \mbox{ otherwise.}
    \end{cases}\label{eq:compat}
\end{equation}

Finally, the probability to select $\bm{a}_t$ is computed with compatibility $u^{*}_j$ using SoftMax function:

\begin{equation} \label{eq:policy_softmax_1}
    p(\bm{a}_t=i|\bm{a}_{1:t-1};\bm{x}) = p_i = \frac{e^{u^{*}_i / \mathcal{T}}}{\sum_j e^{u^{*}_j / \mathcal{T}}} \quad \forall i=1, \ldots, N.
\end{equation}

\subsection{Implementation of SML}
We employ a 2-layer MLP with 128 hidden dimensions to match the scale N with the embedding $\bm{h}$. The sum of these vectors is then input into another 2-layer MLP with ReLU activation to parameterize the SML, $g_{\phi}(\bm{h}, N)$ with parameters $\phi$. For the test time adaptation, we employ a 2-layered MLP with ReLU activation for $g_{\eta}(\bm{h})$, which has 128 hidden dimensions, 128 input dimensions, and 128 output dimensions.

To train the SML, we leverage both distillation and zero-shot simulation objectives. Distillation uses $\{\bm{h}_{(l)}\}_{l=1}^{L}$ where $L=4,000$ and instances $\bm{x}_{(l)}$ uniformly sampled for $N=200, 300, 400$. The zero-shot simulation objective uses the same instances as the distillation process. We only used sparse data with a moderate scale to verify that our method can be trained in limited environments and can be useful in harsher, practical settings.

\subsection{Implementation of SAGE for Target Neural Architectures}

The SAGE algorithm is similar to EAS\citep{hottung2021efficient}, but with different scheduling of locality bias and SoftMax temperature. EAS aims to tune $\bm{h}$, but for memory efficiency during inference, it only focuses on a subset of $\bm{h}$, specifically $\bm{q}_{c}^{*}$. The learnable parameter $\eta$ is trained for this purpose:

\begin{align*}
    \bm{q}_{c}^{*} \leftarrow \bm{q}_{c}^{*} + g_{\eta}(\bm{q}_{c}^{*})
\end{align*}

Where $g_{\eta}$ is a multi-layer perceptron of parameter $\eta$ with ReLU activation. The novel part of SAGE is locality bias and temperature scaling.

\textbf{Locality bias and scheduling.} The locality bias, we utilize distance between current node $x_{\bm{a}_t}$ and candidate nodes:

\begin{equation*} 
    d(x_{\bm{a}_t},\bm{x}) := \left\{||x_i - x_{\bm{a}_t}||; i \in \{1,\ldots,N\} \setminus \bm{a}_{1:t} \right\}.
\end{equation*}
Then we penalize compatibility of high-distance nodes (i.e., compensate local-distance nodes):
\begin{equation} \label{eq:bias}
    u^{*}_{j} \leftarrow u^{*}_{j} - \alpha d(x_{\bm{a}_t},\bm{x})_{j}, \mbox{ } \forall j \in \{1,\ldots,N\} \setminus \bm{a}_{1:t}.
\end{equation}

The reliance of locality bias $\alpha$ is decaying as learning iteration $K$ is increasing:

\begin{equation}
    \alpha \leftarrow \gamma_1 \alpha, \mbox{ where } 0 \leq \gamma_1 < 1.
\end{equation}

\textbf{SoftMax scaling.} The SoftMax temperature $\mathcal{T}$ for \cref{eq:policy_softmax_1} is initially set to a high value for exploration and gradually lowered as training progresses for better exploitation:
\begin{equation}
    \mathcal{T} \leftarrow \gamma_2 \mathcal{T}, \mbox{ where } 0 \leq \gamma_2 < 1.
\end{equation}

\subsection{Hyperparameters for Training}
In the training of the SML model, the ADAM optimizer as described by \citep{adam} is utilized with a learning rate of $10^{-3}$ and a coefficient of $\beta = 1$ for the zero-shot simulation loss $\mathcal{J}_{\text{distil}}(\phi,N)$. Test-time adaptation is performed using the SAGE method, where the initial temperature is set to $\mathcal{T} = 1.0$ and scheduled to decrease to $\mathcal{T} = 0.3$ after $K = 200$ iterations of decay (i.e., $\delta_1^K = 0.3$), where $K$ represents the number of iterations. The initial reliance level of the locality bias is set to $\alpha = 1$ and scheduled to decrease to $\alpha = 0.3$ after $K = 200$ iterations of decay (i.e., $\delta_2^K = 0.3$). The value of $K$ is set to 200 for TSP and CVRP, and 100 for PCTSP and OP. An augmentation technique described by \citep{kwon2020pomo} is implemented to boost solution quality, so we sampled $N \times A$ solutions from the DRL model, where $N$ represents the scale of the problem and $A$ represents the number of augmentations. The number of augmentations, $A$, is adjusted according to the scale of the problem, $N$. The parameter $\eta$ is learned using $\mathcal{J}_{\text{SAGE}}$ and the final solution is chosen as the best solution among the $K \times N \times A$ solutions. Test-time adaptation is performed using the ADAM optimizer with a learning rate as specified in the open source code by \citep{hottung2021efficient}. The weight of the self-imitation learning objective is set to $\lambda = 0.005$.

\subsection{Instance generation and evaluation metric}

\textbf{Instance generation.} We generate training and test instances following \citet{kool2018attention}, like many works in NCO \citep{kwon2020pomo, kim2022sym, hottung2021efficient}.

\begin{itemize}
    \item TSP: ($x, y$) coordination of each node is randomly sampled to obtain a uniform distribution within the range of $[0,1]$.
    \item CVRP: ($x, y$) coordination of each node is uniformly distributed within $[0,1]$, while the demand of each node is sampled from the uniform distribution within the range of $[1, 9]$.
    \item PCTSP: ($x, y$) coordination and the penalty of each node is sampled from the uniform distribution within the range of $[0,1]$.
    \item OP: ($x, y$) coordination of each node is randomly sampled to obtain a uniform distribution within $[0,1]$. Furthermore, each node's price ($\rho_i$) is determined based on its distance to the depot ($\rho_i = 1 +\lfloor 99 \cdot \frac{d_{0i}}{max_{j=1}^{n}d_{0j}}\rfloor$), such that it increases with distance.
\end{itemize}

\textbf{Evaluation metric.} We measure the performance using baseline solvers for each task. Note that this is not an optimality gap since these baselines are not exact methods except for Concorde, which is based on the cutting plane method. Precisely, (Performance) Gap is computed as 
$$ \text{Performance Gap} = \frac{obj - obj_B}{obj_B} \times 100 (\%),$$
where $obj$ is the objective value of DRL methods, and $obj_B$ is the objective value from baseline solvers (Concode, LKH3, OR-Tools, and Compass).

\clearpage

\section{Implementation Details of Baselines} \label{append:imp_detail_baseline}

\subsection{Active Search}
Active Search is tuning all parameters of a model to a target instance. For TSP and CVRP, we follow the learning rate hyperparameter instruction of open source code by \citep{hottung2021efficient}.

\begin{table}[h]
    \centering
    \begin{tabular}{lcccc}
    \hline
    &TSP & CVRP &PCTSP &OP\\
        \hline
        Learning rate & 2.6e-4 & 2.6e-5 & 2.6e-4 & 2.6e-4 \\
        \hline
    \end{tabular}
    \caption{Hyperparameter setting for active search.}
\end{table}

\subsection{Efficient Active Search}
Efficient Active Search(EAS) inserts the layers into the model and updates the layers of parameters. We follow the position of inserted layer and hyperparameter, same with open source code by \citep{hottung2021efficient} except the imitation rate, batch size, and augmentation size, which are the same as our method.

\begin{table}[h]
    \centering
    \begin{tabular}{lcccc}
    \hline
    &TSP & CVRP &PCTSP &OP\\
        \hline
        Learning rate & 3.2e-3 & 4.1e-3 & 1e-3 & 1e-3 \\ 
        Imitation rate & 5e-3 & 5e-3 & 5e-3 & 5e-3 \\
        \hline
    \end{tabular}
    \caption{Hyperparameter setting for efficient active search.}
\end{table}

\subsection{Simulation Guided Beam Search with EAS}

Simulation Guided Beam Search(SGBS) with EAS is combined SGBS with EAS. We follow the position of inserted layer and hyperparameter, same with open source code by \citep{choo2022simulation} except the batch size, and augmentation size, which are the same as our method.

\begin{table}[h]
    \centering
    \begin{tabular}{lcc}
    \hline
    &TSP & CVRP\\
        \hline
        Learning rate & 8.15e-3 & 4.1e-3\\
        Imitation rate & 6e-3 & 5e-3\\
        Beam width & 10 & 4\\
        Rollout per node & 9 & 3\\
        EAS iteration & 1 & 1\\
        \hline
    \end{tabular}
    \caption{Hyperparameter setting for simulation guided beam search with eas.}
\end{table}

\clearpage

\section{Additional Experiments}
\label{append:addional-exp}

\subsection{Adaptation Performance of TSP and OP} 
\label{adap-tsp-op}
\begin{figure*}[!ht]
     \centering
     \begin{subfigure}[b]{0.32\textwidth}
         \centering
         \includegraphics[width=\textwidth]{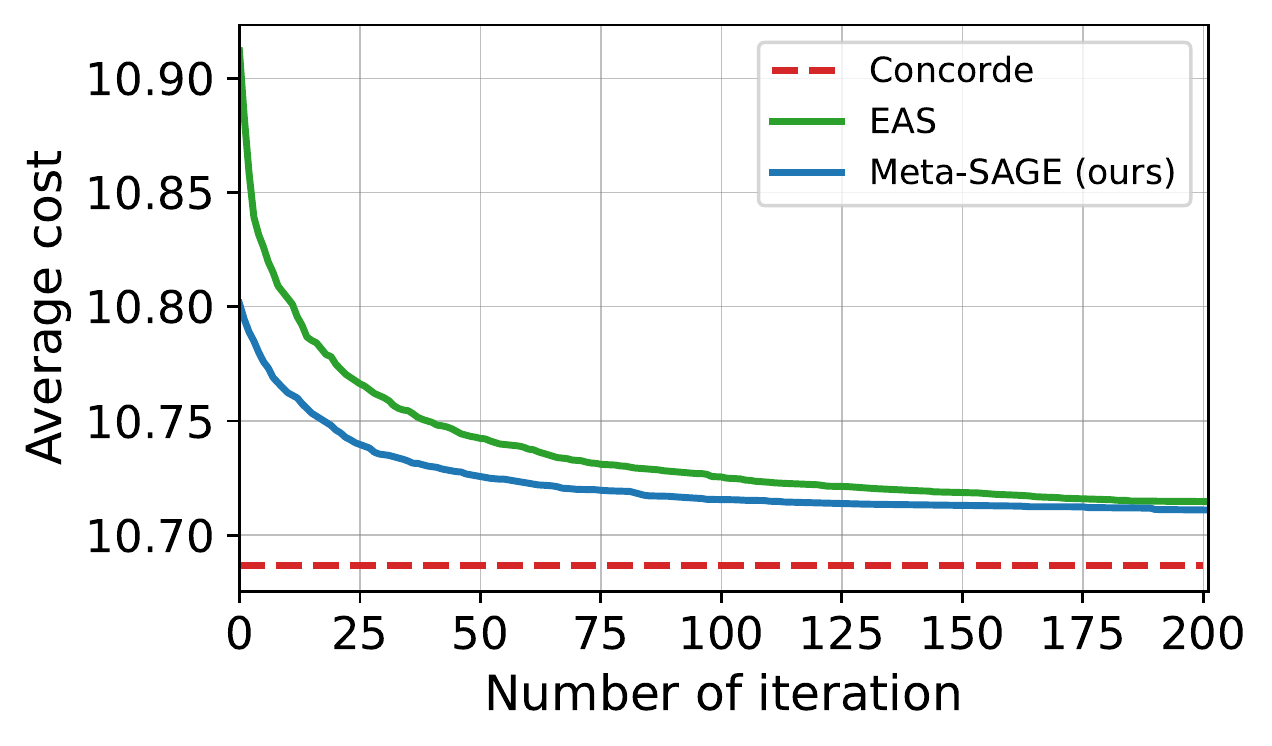}
         \caption{$N=200$}
         \label{fig:perf_shot_tsp_200}
     \end{subfigure}
     \begin{subfigure}[b]{0.32\textwidth}
         \centering
        \includegraphics[width=\textwidth]{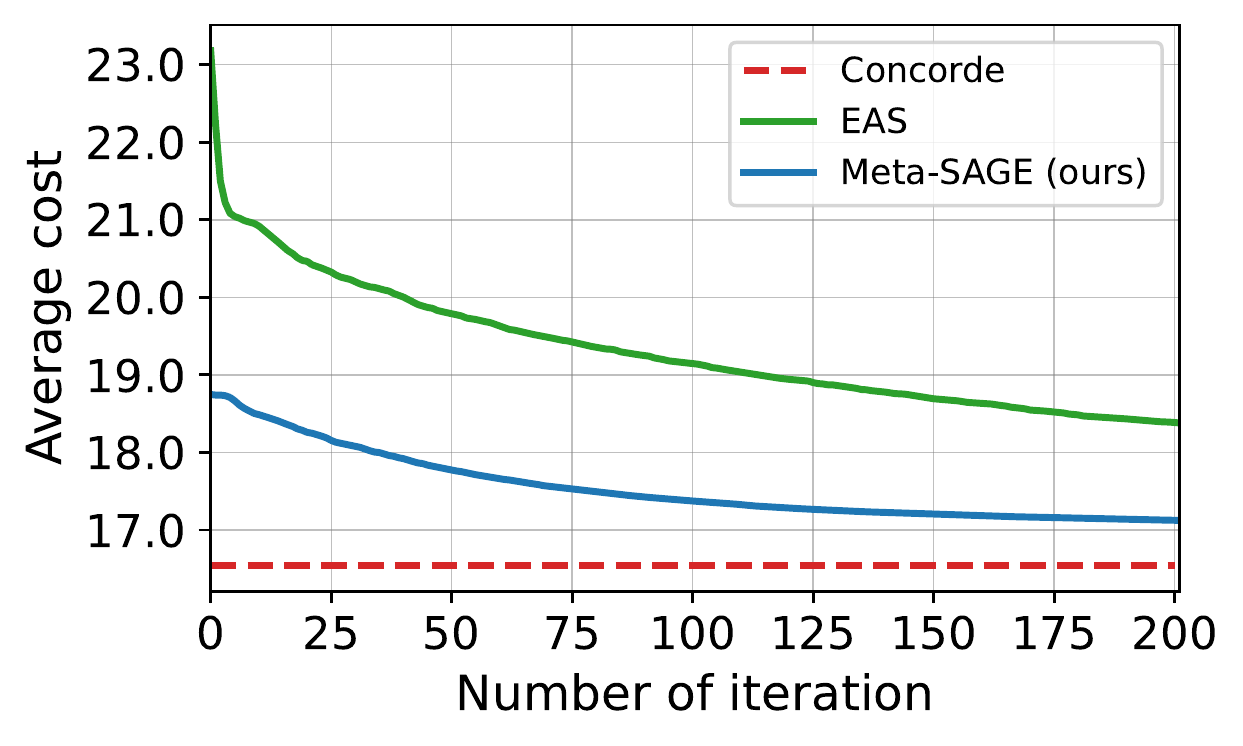}
         \caption{$N=500$}
         \label{fig:perf_shot_tsp_500}
     \end{subfigure}
          \begin{subfigure}[b]{0.32\textwidth}
         \centering
        \includegraphics[width=\textwidth]{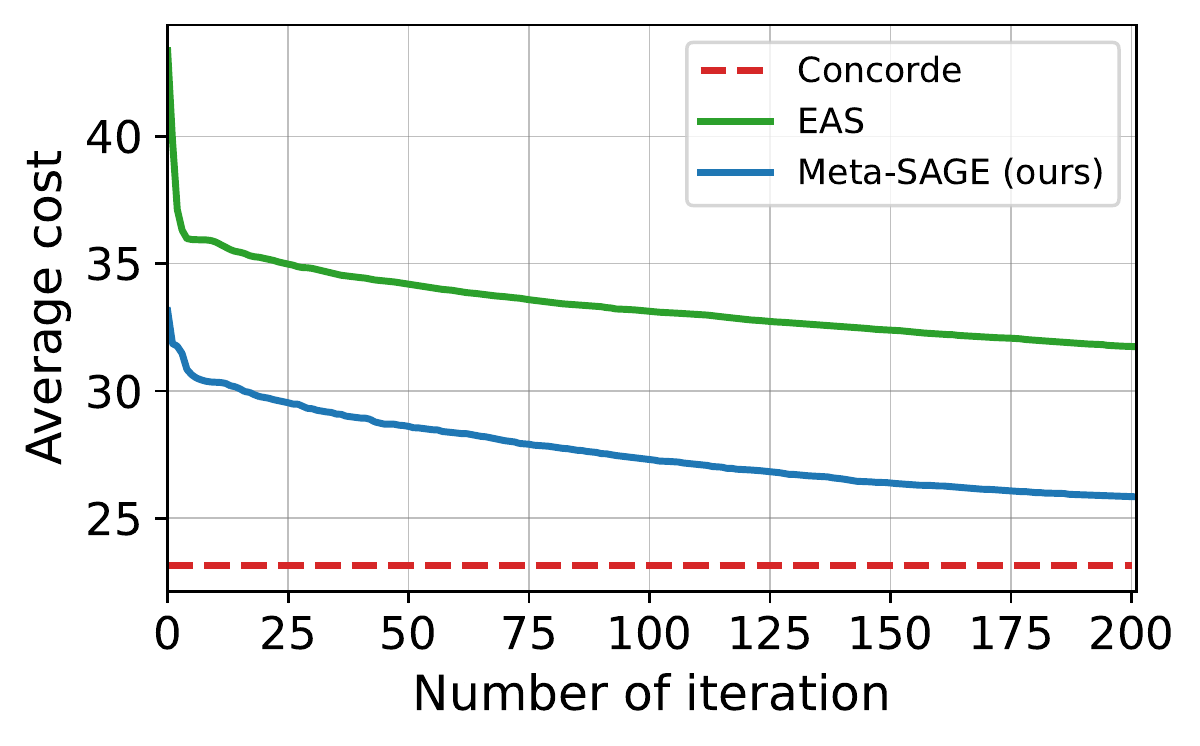}
         \caption{$N=1,000$}
         \label{fig:perf_shot_tsp_1000}
     \end{subfigure}
     \caption{Adaptation performance on TSP with pre-trained Sym-NCO compared to Concorde.}
    \label{fig:perf_shot_tsp}
\end{figure*}

\begin{figure*}[!ht]
     \centering
     \begin{subfigure}[b]{0.32\textwidth}
         \centering
         \includegraphics[width=\textwidth]{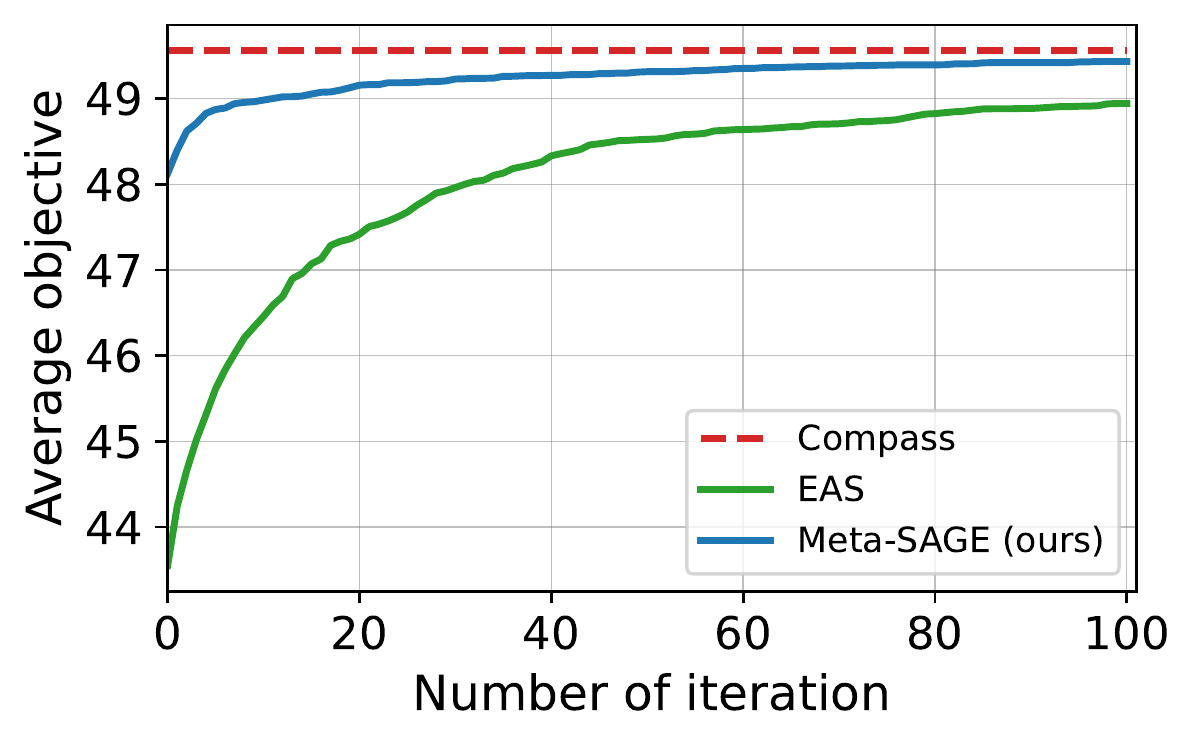}
         \caption{$N=200$}
         \label{fig:perf_shot_op_200}
     \end{subfigure}
     \begin{subfigure}[b]{0.32\textwidth}
         \centering
        \includegraphics[width=\textwidth]{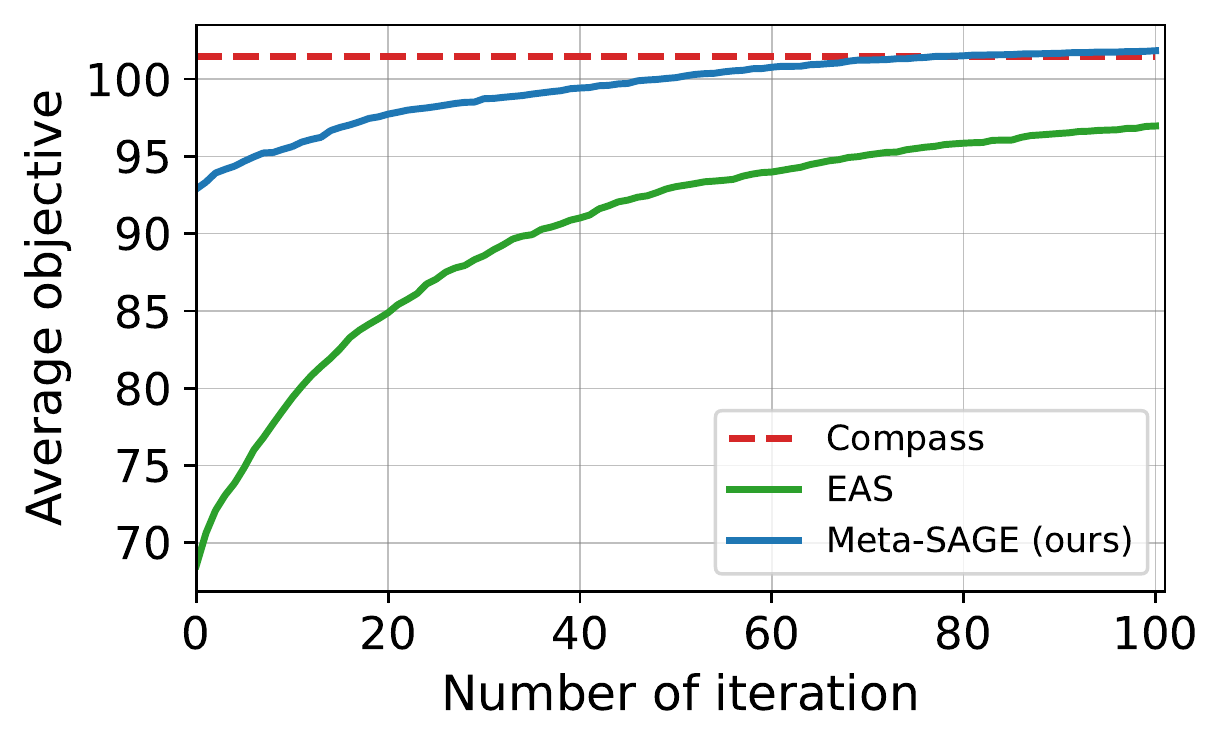}
         \caption{$N=500$}
         \label{fig:perf_shot_op_500}
     \end{subfigure}
          \begin{subfigure}[b]{0.32\textwidth}
         \centering
        \includegraphics[width=\textwidth]{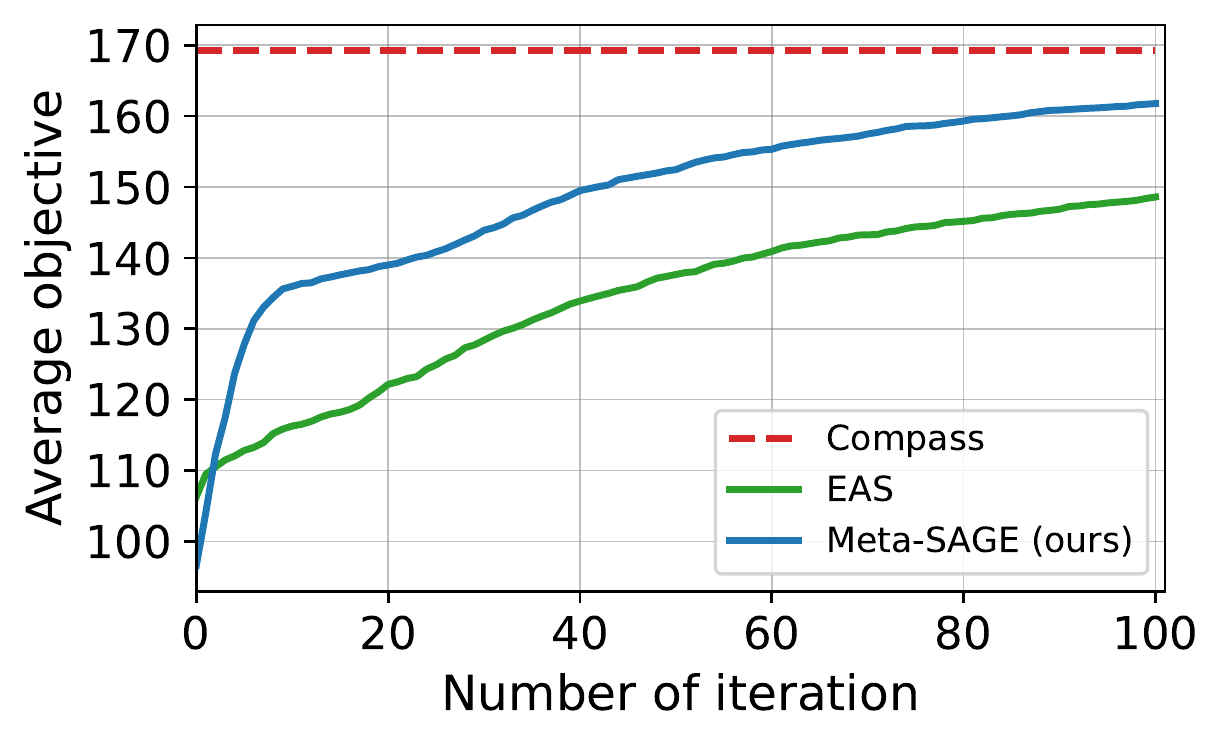}
         \caption{$N=1,000$}
         \label{fig:perf_shot_op_1000}
     \end{subfigure}
     \caption{Adaptation performance on OP with pre-trained Sym-NCO compared to Compass.}
    \label{fig:perf_shot_op}
\end{figure*}

\cref{fig:perf_shot_tsp} and \cref{fig:perf_shot_op} depict the cost obtained by the EAS method and our proposed method with respect to the number of adaptation iterations on TSP and OP tasks, respectively. The results demonstrate that our method enhances the performance across all scales and tasks with the same number of iterations compared to EAS. Specifically, for the OP task with $N = 500$, our method outperforms Compass, a problem-specific solver, with 100 iterative adaptations.

\label{append:time-performance}

\subsection{Full results for X-instances in CVRPLIB}
\label{append:cvrplib_full}
The dataset consists of $100$ instances with a scale range from $100$ to $1,000$. We provide the full experimental result below.
\begin{table}[!ht]
\centering
\caption{Evaluation of adaptation methods on the real-world CVRP dataset. We report average values according to the range of instance scale $N$.} \label{cvrplib_rst_full}
\scalebox{0.9}{\begin{tabular}{cccc}
\toprule[1.0pt]
\multirowcell{2}[-0.7ex]{Range} &\multirowcell{2}[-0.7ex]{LKH3} & \multicolumn{2}{c}{Sym-NCO}
\\
\cmidrule[0.5pt](lr{0.2em}){3-4}
 & & EAS & Ours 
\\
\specialrule{0.5pt}{2pt}{4pt}
$100 \leq N < 200$ &25,827&26,196& \textbf{26,169}\\
$200 \leq N < 300$ &41,956&42,553& \textbf{42,392}\\
$300 \leq N < 400$ &56,588&57,675& \textbf{57,245}\\
$400 \leq N < 500$ &76,087&78,316& \textbf{77,238}\\
$500 \leq N < 600$ &92,095&95,384& \textbf{94,028}\\
$600 \leq N < 700$ &85,453&88,987& \textbf{87,372}\\
$700 \leq N < 800$ &89,009&92,288& \textbf{90,184}\\
$800 \leq N < 900$ &112,785&119,332& \textbf{116,096}\\
$900 \leq N \leq 1000$&150,332&175,999& \textbf{172,916}\\

\bottomrule[1.0pt]
\end{tabular}}
\end{table}

\clearpage

\subsection{Effectiveness of Objective Function Components in Training SML}
\label{append:distil_ablation}

\begin{figure*}[!ht]
     \centering
     \begin{subfigure}[b]{0.335\textwidth}
         \centering
         \includegraphics[width=\textwidth]{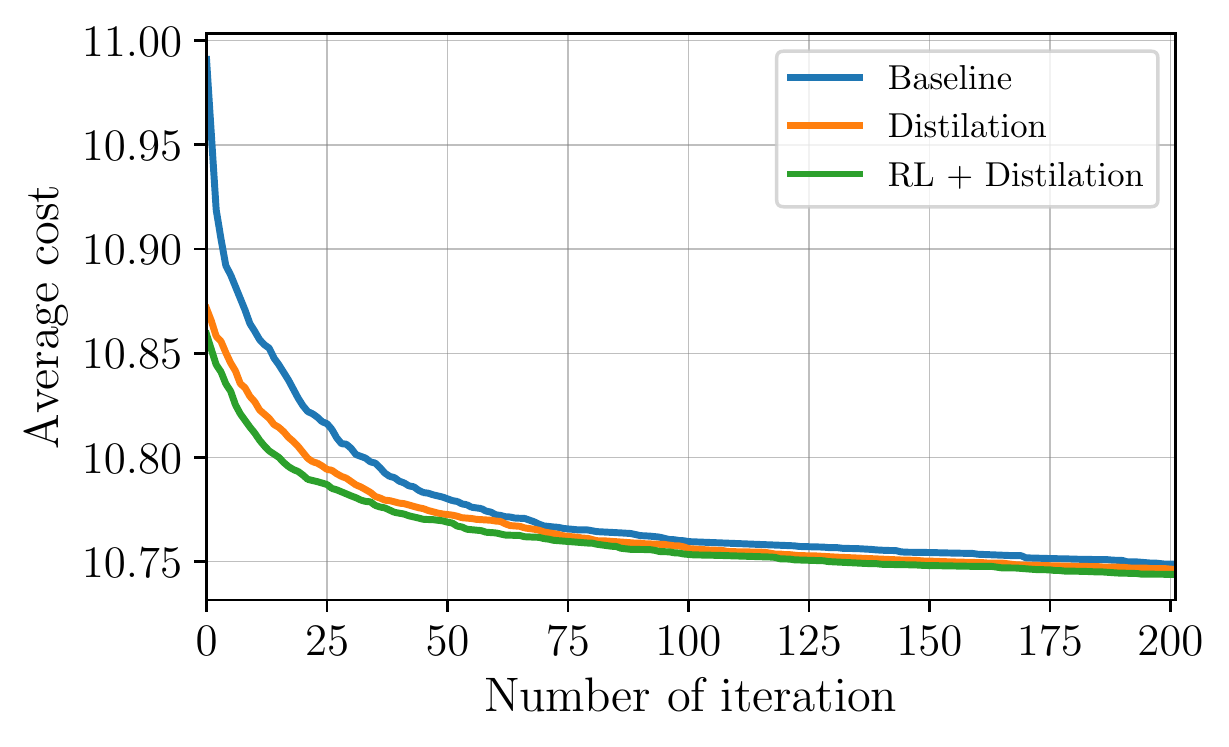}
         \caption{$N=200$}
     \end{subfigure}
     \begin{subfigure}[b]{0.32\textwidth}
         \centering
        \includegraphics[width=\textwidth]{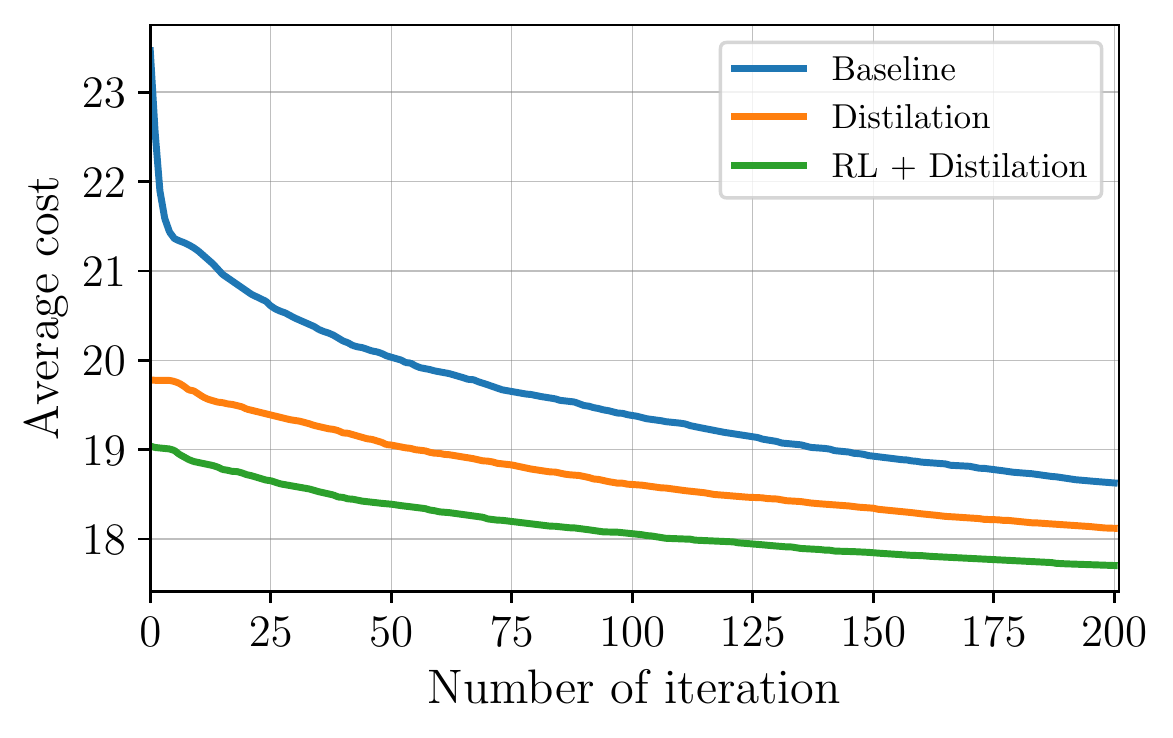}
         \caption{$N=500$}
     \end{subfigure}
          \begin{subfigure}[b]{0.32\textwidth}
         \centering
        \includegraphics[width=\textwidth]{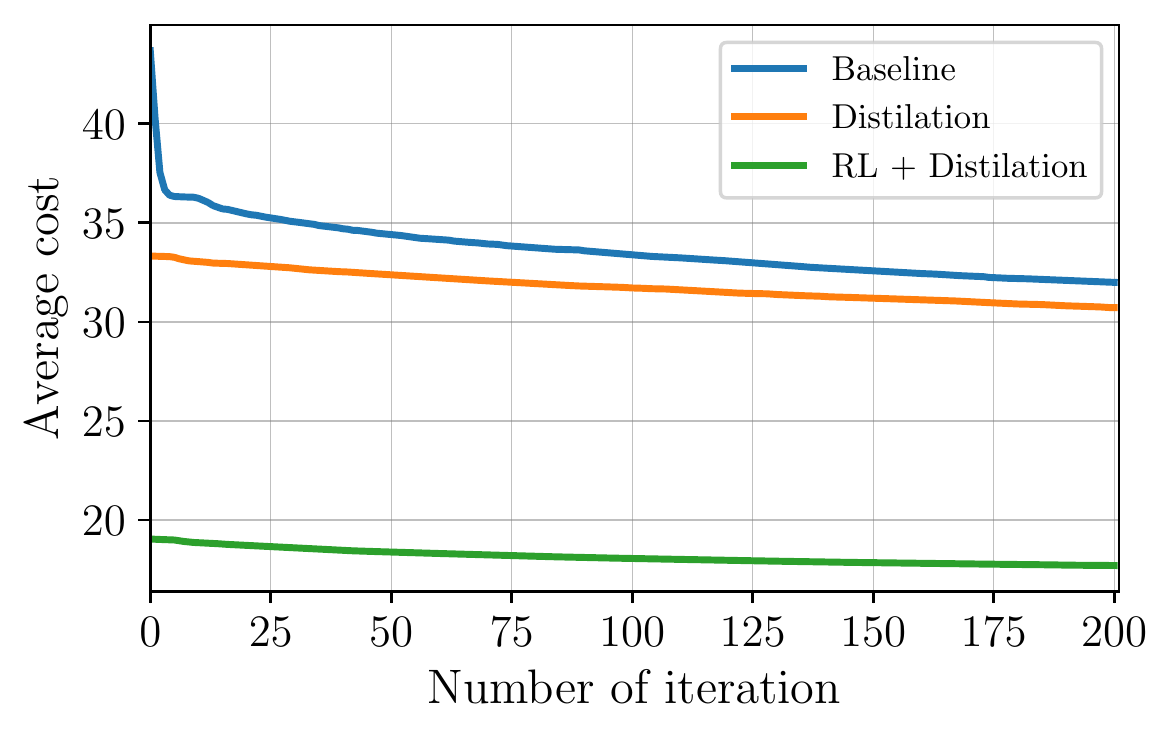}
         \caption{$N=1,000$}
     \end{subfigure}\label{distil_abl}
     \caption{Ablation study for SML componets}
\end{figure*}

We evaluated the effectiveness of each component of the distillation objective and RL objective (i.e., $\mathcal{J}_{\text{zero}}$) by analyzing the performance improvements. The results in \cref{distil_abl} clearly show that each component contributed to improved performance. The effectiveness increased as the scale $N$ increased, indicating that our SML successfully mitigates distributional shifts on a larger scale. Additionally, $\mathcal{J}_{\text{zero}}$, which is designed to increase zero-shot capability, was verified to significantly improve the performance in zero-shot scenarios (i.e., $K=0$), especially on $N=1000$.

\clearpage

\end{document}